\documentclass[nohyperref]{article}

\usepackage{microtype}
\usepackage{graphicx}
\usepackage{subfigure}
\usepackage{booktabs} %
\usepackage{multirow}
\usepackage{threeparttable}

\usepackage{hyperref}

\usepackage{times}
\usepackage{latexsym}
\usepackage[T1]{fontenc}
\usepackage[utf8]{inputenc}

\usepackage[accepted]{icml2022}

\usepackage{amsmath}
\usepackage{amssymb}
\usepackage{mathtools}
\usepackage{amsthm}
\usepackage{enumitem}
\usepackage{xspace}
\usepackage{soul}

\usepackage[capitalize,noabbrev]{cleveref}

\theoremstyle{plain}

\theoremstyle{definition}

\theoremstyle{remark}

\usepackage{caption}

\newcommand{\name}{FLASH\xspace}
\newcommand{\qname}{FLASH-Quad\xspace}
\newcommand{\lname}{FLASH\xspace}
\newcommand{\attn}{mixed chunk attention\xspace}

\usepackage[textsize=tiny]{todonotes}

\usepackage{listings}
\usepackage{textcomp}

\newfloat{Code}{htbp}{Code}

\newcommand{\tablestyle}[2]{\setlength{\tabcolsep}{#1}\renewcommand{\arraystretch}{#2}\centering\scriptsize}

\icmltitlerunning{}

\begin{document}

\twocolumn[
\icmltitle{Transformer Quality in Linear Time}

\icmlsetsymbol{equal}{*}

\begin{icmlauthorlist}
\icmlauthor{Weizhe Hua}{equal,yyy,comp}
\icmlauthor{Zihang Dai}{equal,comp}
\icmlauthor{Hanxiao Liu}{equal,comp}
\icmlauthor{Quoc V. Le}{comp}
\end{icmlauthorlist}

\icmlaffiliation{yyy}{Cornell University}
\icmlaffiliation{comp}{Google Research, Brain Team}

\icmlcorrespondingauthor{Weizhe Hua}{wh399@cornell.edu}
\icmlcorrespondingauthor{Zihang Dai}{zihangd@google.com}
\icmlcorrespondingauthor{Hanxiao Liu}{hanxiaol@google.com}
\icmlkeywords{Machine Learning, ICML}

\vskip 0.3in
]

\printAffiliationsAndNotice{\icmlEqualContribution} %

\setcounter{footnote}{2}

\begin{abstract}
We revisit the design choices in Transformers, and propose methods to address their weaknesses in handling long sequences. First, we propose a simple layer named gated attention unit, which allows the use of a weaker 
single-head attention with minimal quality loss.
We then propose a linear approximation method complementary to this new layer, which is accelerator-friendly and highly competitive in quality.
The resulting model, named \name\footnotemark, matches the perplexity of improved Transformers over both short (512) and long (8K) context lengths, achieving training speedups of up to 4.9$\times$ on Wiki-40B and 12.1$\times$ on PG-19 for auto-regressive language modeling, and 4.8$\times$ on C4 for masked language modeling.

\end{abstract}

\section{Introduction}
\label{sec:intro}
Transformers~\cite{vaswani2017attention} have become the new engine of state-of-the-art deep learning systems, leading to many recent breakthroughs in language~\cite{devlin2018bert, brown2020language} and vision~\cite{dosovitskiy2020image}. Although they have been growing in model size, most Transformers are still 
limited to short context size due to their quadratic complexity over the input length. %
This limitation prevents Transformer models from processing long-term information, a critical property for many applications.

Many techniques have been proposed to speedup Transformers over extended context via more efficient attention mechanisms~\cite{child2019generating, dai2019transformer, rae2019compressive, choromanski2020rethinking, wang2020linformer, katharopoulos2020transformers, beltagy2020longformer, zaheer2020big, kitaev2020reformer, roy2021efficient, jaegle2021perceiver}.
Despite the linear theoretical complexity for some of those methods, vanilla Transformers still remain as the dominant choice in state-of-the-art systems.
Here we examine this issue from a practical perspective, and find existing efficient attention methods suffer from at least one of the following drawbacks:

\begin{figure}[t]
\centering
\includegraphics[width=0.95\linewidth]{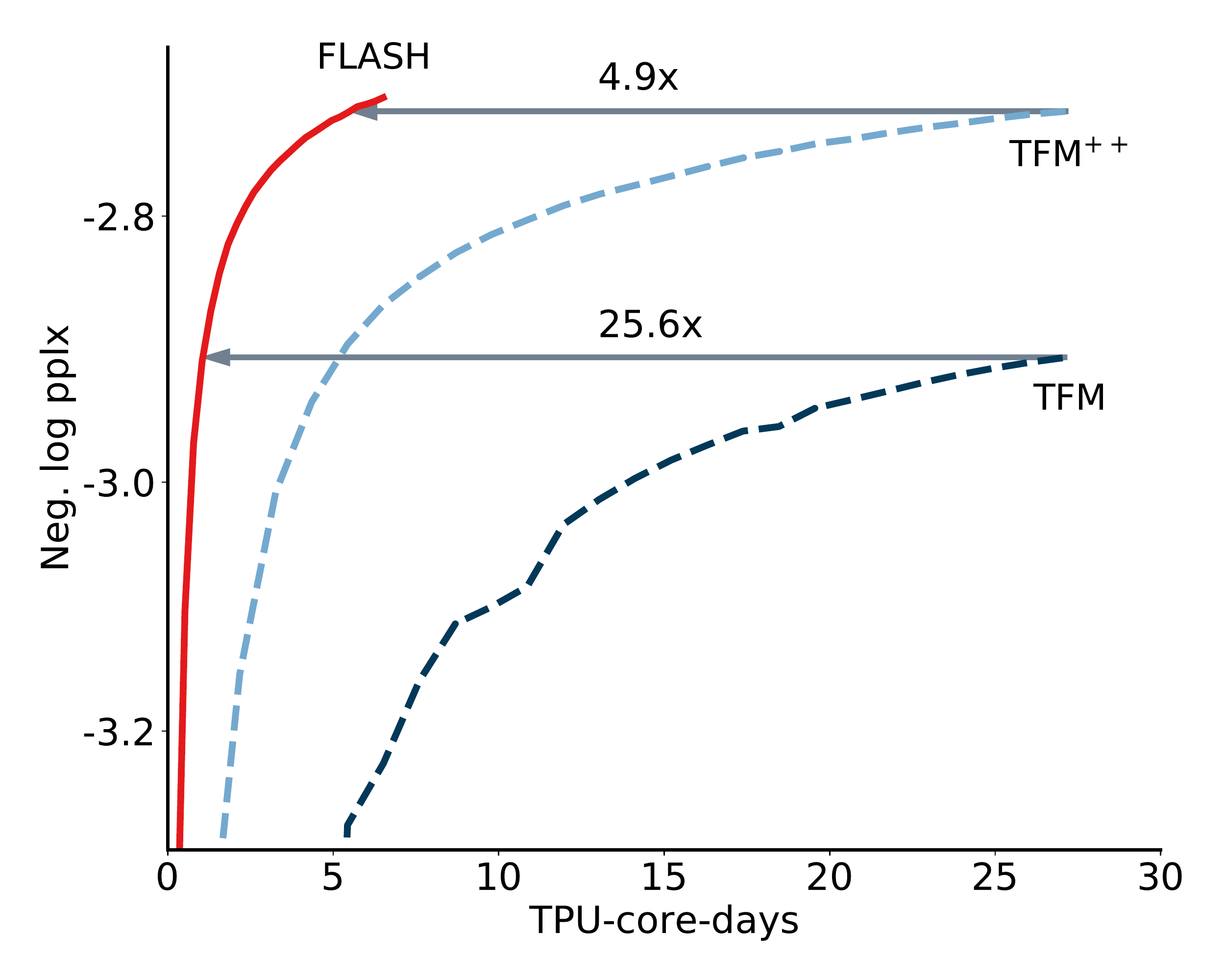}
    \hspace{-40mm}\resizebox{.45\columnwidth}{!}{\tablestyle{2pt}{1}
	\begin{tabular}[b]{c c c}
		& \multicolumn{2}{c}{Speedup} \\
		Length & over TFM & over TFM++ \\
		\cmidrule{1-1} \cmidrule(lr){2-2} \cmidrule{3-3}
		512  & 1.8$\times$ & 1.2$\times$ \\
		1024 & 9.0$\times$ & 1.3$\times$ \\
		2048 & 8.9$\times$ & 1.6$\times$ \\
		4096 & 13.1$\times$ & 2.7$\times$ \\
		8192 & \textbf{25.6$\times$} & \textbf{4.9$\times$} \\
	\vspace{7mm}
	\end{tabular}}
\caption{TPU-v4 training speedup of \lname relative to the vanilla Transformer (TFM) and an augmented Transformer (TFM++) for auto-regressive language modeling on Wiki-40B --- \small{
All models are comparable in size at around 110M and trained for 125K steps with 2$^{18}$ tokens per batch.}
}
\label{fig:speedup}
\end{figure}

\begin{figure*}
\begin{minipage}{0.67\linewidth}
    \includegraphics[width=\linewidth]{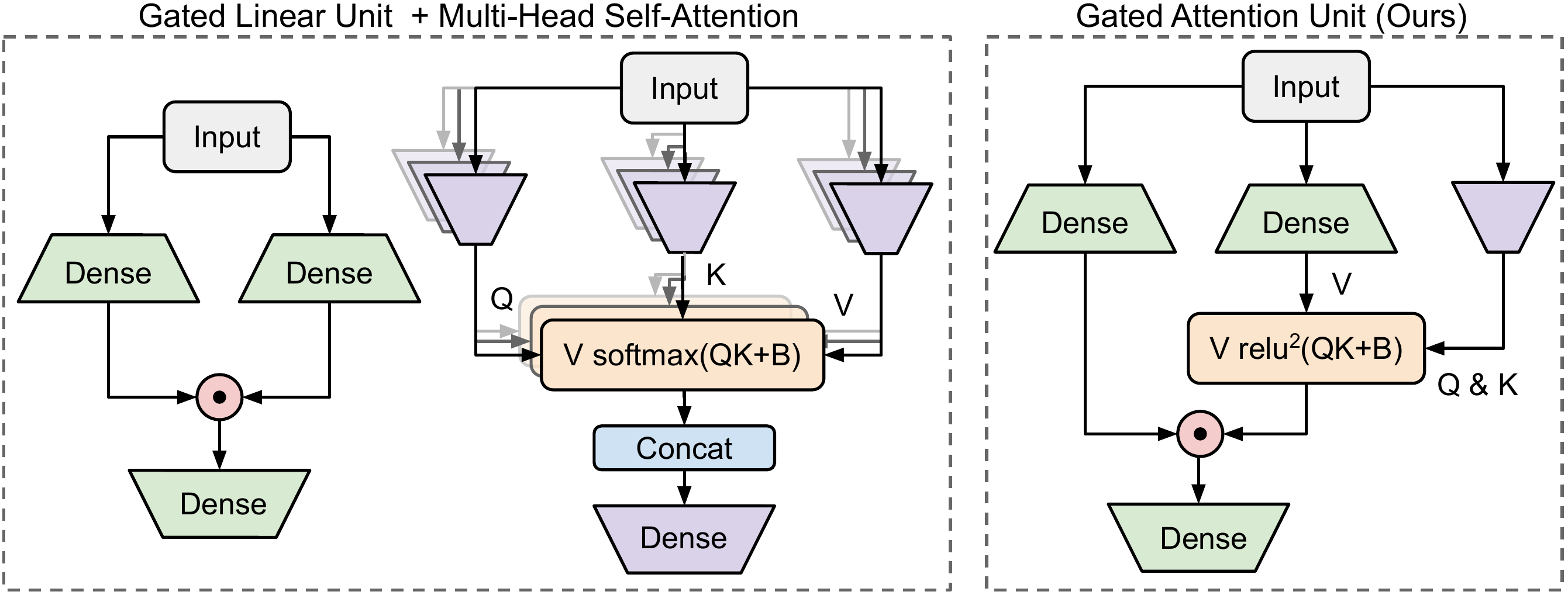}
\end{minipage}
\hfill
\begin{minipage}{0.31\linewidth}
\begin{lstlisting}[language=python,mathescape]
def scale_offset(x):
    gamma = var(x.shape[-1:])
    beta = var(x.shape[-1:])
    return x * gamma + beta
    
def attn(x, v, s=128):
    z = dense(x, s)
    q, k = scale_offset(z), scale_offset(z)
    qk = tf.einsum('bns,bms$\rightarrow$bnm', q, k)
    a = relu(qk + rel_pos_bias(q, k)) ** 2
    return tf.einsum('bnm,bme$\rightarrow$bne', a, v)
    
def gated_attn_unit(x, d=768, e=1536):
    shortcut, x = x, norm(x)
    u, v = dense(x, e), dense(x, e)
    x = u * attn(x, v)
    return dense(x, d) + shortcut
\end{lstlisting}
\label{code:gau}
\end{minipage}
\caption{(a) An augmented Transformer layer which consists of two blocks: Gated Linear Unit (GLU) and Multi-Head Self-Attention (MHSA), (b) Our proposed Gated Attention Unit (GAU), (c) Pseudocode for Gated Attention Unit. Skip connection and input normalization over the residual branch are omitted in (a), (b) for brevity.}
\label{fig:layers}
\end{figure*}

\footnotetext{FLASH = \textbf{F}ast \textbf{L}inear \textbf{A}ttention with a \textbf{S}ingle \textbf{H}ead}
\begin{itemize}[leftmargin=*,topsep=0pt,itemsep=0pt]
\item \textbf{Inferior Quality}. Our studies reveal that vanilla Transformers, when augmented with several simple tweaks, can be much stronger than the common baselines used in the literature (see Transformer vs. Transformer++ in Figure~\ref{fig:speedup}). Existing efficient attention methods often incur significant quality drop compared to augmented Transformers, and this drop outweighs their efficiency benefits. 

\item \textbf{Overhead in Practice}.
As efficient attention methods often complicate Transformer layers and require extensive memory re-formatting operations,
there can be a nontrivial gap between their theoretical complexity and empirical speed on accelerators such as GPUs or TPUs.

\item \textbf{Inefficient Auto-regressive Training}. Most attention linearization techniques enjoy fast decoding during inference, but can be extremely slow to train on auto-regressive tasks such as language modeling.
This is primarily due to their RNN-style sequential state updates over a large number of steps, making it infeasible to fully leverage the strength of modern accelerators during training.
\end{itemize}

We address the above issues by developing a new model family that, for the first time, not only achieves parity with fully augmented Transformers in quality, but also truly enjoys linear scalability over the context size on modern accelerators.
Unlike existing efficient attention methods which directly aim to approximate the multi-head self-attention (MHSA) in Transformers,
we start with a new layer design which naturally enables higher-quality approximation.
Specifically, our model, named \name, is developed in two steps:

First, we propose a new layer that is more desirable for effective approximation.
We introduce a gating mechanism to alleviate the burden of self-attention, resulting in the \emph{Gated Attention Unit} (GAU) in Figure~\ref{fig:layers}.
As compared to Transformer layers, each GAU layer is cheaper, and more importantly, its quality relies less on the precision of attention. In fact, GAU with a small single-head, softmax-free attention is as performant as Transformers. While GAU still suffers from quadratic complexity over the context size, it weakens the role of attention hence allows us to carry out approximation later with minimal quality loss.

We then propose an efficient method to approximate the quadratic attention in GAU,
leading to a layer variant with linear complexity over the context size. The key idea is to first group tokens into chunks, then using precise quadratic attention within a chunk and fast linear attention across chunks, as illustrated in Figure~\ref{fig:linear-attention}.
We further describe how an accelerator-efficient implementation can be naturally derived from this formulation,
achieving linear scalability in practice with only a few lines of code change.

\begin{figure*}[t]
\centering
\footnotesize{
\begin{tabular}[t]{l@{\hskip 1.5em}ccc}
\cmidrule[\heavyrulewidth]{2-4}
    \multirow{10}{*}{\includegraphics[width=0.6\linewidth, height=4.6cm]{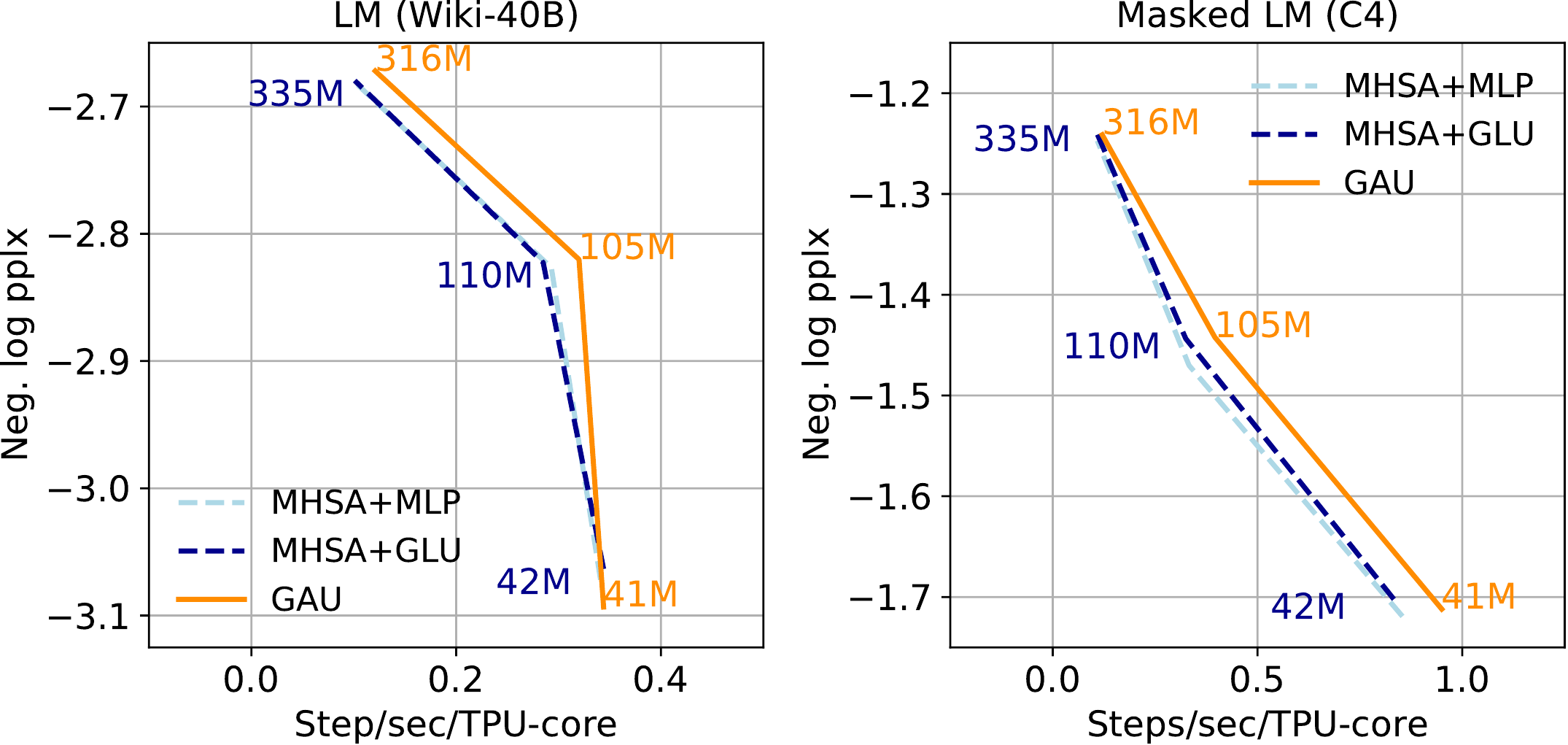}}
    & Layer Type & \# of Layers & $d$ \\\cmidrule{2-4}
    & \multirow{3}{*}{MHSA+MLP}
    & 8+8 & 512 \\
    & & 12+12 & 768  \\
    & & 24+24 & 1024 \\ \cmidrule{2-4}
    & \multirow{3}{*}{MHSA+GLU}
    & 8+8 & 512 \\
    & & 12+12 & 768  \\
    & & 24+24 & 1024 \\ \cmidrule{2-4}
    & \multirow{3}{*}{GAU} & 15 & 512 \\
    & & 22 & 768 \\ 
    & & 46 & 1024 \\
    \cmidrule[\heavyrulewidth]{2-4}
    \end{tabular}
    \vspace{0.55cm}
    \caption{GAU vs. Transformers for auto-regressive and masked language modeling on short context length (512).}
    \label{fig:gau-vs-transformer}
}
\end{figure*}

We conduct extensive experiments to demonstrate the efficacy of \name over a variety of tasks (masked and auto-regressive language modeling), datasets (C4, Wiki-40B, PG-19) and model scales (110M to 500M).
Remarkably,
\name is competitive with fully-augmented Transformers (Transformer++) in quality across a wide range of context sizes of practical interest (512--8K),
while achieving linear scalability on modern hardware accelerators. 
For example, with comparable quality,
\name achieves a speedup of 1.2$\times$--4.9$\times$ for language modeling on Wiki-40B and a speedup of 1.0$\times$--4.8$\times$ for masked language modeling on C4 over Transformer++.
As we further scale up to PG-19~\cite{rae2019compressive}, \name reduces the training cost of Transformer++ by up to 12.1$\times$ and achieves significant gain in quality.

\section{Gated Attention Unit}
\label{sec:quad}
Here we present Gated Attention Unit (GAU), a simpler yet more performant layer than Transformers.
While GAU still has quadratic complexity over the context length, it is more desirable for the approximation method to be presented in Section~\ref{sec:approx-attn}.
We start with introducing related layers:

\paragraph{Vanilla MLP.} Let $X \in \mathbb{R}^{T \times d}$ be the representations over $T$ tokens. The output for Transformer's MLP can be formulated as $O = \phi(XW_u) W_o$ where $W_u \in \mathbb{R}^{d \times e}$, $W_o \in \mathbb{R}^{e \times d}$. Here $d$ denotes the model size, $e$ denotes the expanded intermediate size, and $\phi$ is an element-wise activation function.

\paragraph{Gated Linear Unit (GLU).} This is an improved MLP augmented with gating~\cite{dauphin2017glu}. GLU has been proven effective in many cases~\cite{shazeer2020glu, narang2021transformer} and is used in state-of-the-art Transformer language models~\cite{du2021glam, thoppilan2022lamda}.
\begin{align}
U &= \phi_u(XW_u), \quad V = \phi_v(XW_v) &&\in \mathbb{R}^{T \times e} \label{eq:glu-uv} \\
O &= (U \odot V)W_o &&\in \mathbb{R}^{T \times d} \label{eq:glu-gate}
\end{align}
where $\odot$ stands for element-wise multiplication.
In GLU, each representation $u_i$ is gated by another representation $v_i$ associated with the same token.

\paragraph{Gated Attention Unit (GAU).}
The key idea is to formulate attention and GLU as a unified layer and to share their computation as much as possible (Figure~\ref{fig:layers}).
This not only results in higher param/compute efficiency,
but also naturally enables a powerful attentive gating mechanism.
Specifically,
GAU generalizes Eq.~\eqref{eq:glu-gate} in GLU as follows:
\begin{align}
\label{eq:gau-gating}
  O &= (U \odot \hat{V})W_o \quad \text{where} \quad \hat{V} = A V
\end{align}
where $A \in \mathbb{R}^{T \times T}$ contains token-token attention weights.
Unlike GLU which always uses $v_i$ to gate $u_i$ (both associated with the same token), our GAU replaces $v_i$ with a potentially more relevant representation $\hat{v}_i = \sum_j a_{ij} v_j$ ``retrieved'' from all available tokens using attention. The above will reduce to GLU when $A$ is an identity matrix.

Consistent with the findings in~\citet{liu2021pay},
the presence of gating allows the use of a much simpler/weaker attention mechanism than MHSA without quality loss:
\begin{align}
\label{eq:gau-key}
  Z &= \phi_z(X W_z) &&\in \mathbb{R}^{T \times s} \\
A &= \mathrm{relu}^2\left(\mathcal{Q}(Z) \mathcal{K}(Z)^\top + b\right) &&\in \mathbb{R}^{T \times T}
\end{align}

\begin{table}[!ht]
\small
\begin{tabular}{@{}l|cc@{}}
\toprule
Modifications & PPLX (LM/MLM) & Params (M) \\ \midrule
original GAU & \textbf{16.78} / \textbf{4.23} & 105 \\ \midrule
relu$^2$ $\xrightarrow{}$ softmax & 17.04 / 4.31 & 105 \\
single-head $\xrightarrow{}$ multi-head & 17.76 / 4.48 & 105 \\
no gating & 17.45 / 4.58 & 131 \\ \bottomrule
\end{tabular}
\caption{Impact of various modifications on GAU.}
\label{tab:gau-ablation}
\end{table}

\begin{table}[!ht]
\small
\begin{tabular}{@{}l|cc@{}}
\toprule
Modifications & PPLX (LM/MLM) & Params (M) \\ \midrule
original MHSA & \textbf{16.87} / \textbf{4.35} & 110 \\ \midrule
softmax $\xrightarrow{}$ relu$^2$ & 17.15 / 4.77 & 110 \\
multi-head $\xrightarrow{}$ single-head & 17.89 / 4.73 & 110 \\
add gating & 17.25 / 4.43 & 106 \\ \bottomrule
\end{tabular}
\caption{Impact of various modifications on MHSA.}
\label{tab:mhsa-ablation}
\end{table}

where $Z$ is a shared representation ($s \ll d$)\footnote{Unless otherwise specified, we set $s=$128 in this work.},
$\mathcal{Q}$ and $\mathcal{K}$ are two cheap transformations that apply per-dim scalars and offsets to $Z$ (similar to the learnable variables in LayerNorms),
and $b$ is the relative position bias.
We also find the softmax in MHSA can be simplified as a regular activation function in the case of GAU\footnote{We use squared ReLU~\cite{so2021primer} throughout this paper, which empirically works well on language tasks.}.
The GAU layer and its pseudocode are illustrated in Figure~\ref{fig:layers}.

Unlike Transformer's MHSA which comes with $4 d^2$ parameters,
GAU's attention introduces only a single small dense matrix $W_z$ with $ds$ parameters on top of GLU (scalars and offsets in $\mathcal{Q}$ and $\mathcal{K}$ are negligible).
By setting $e = 2d$ for GAU,
this compact design allows us to replace each Transformer block (MLP/GLU + MHSA) with two GAUs while retaining similar model size and training speed.

\paragraph{GAU vs. Transformers.} Figure~\ref{fig:gau-vs-transformer} shows that GAUs are competitive with Transformers (MSHA + MLP/GLU) on TPUs across different models sizes.
Note these experiments are conducted over a relatively short context size (512). We will see later in Section~\ref{sec:experiment} that GAUs are in fact even more performant when the context length is longer, thanks to their reduced capacity in attention.

\paragraph{Layer Ablations.} 
In Table~\ref{tab:gau-ablation} \&~\ref{tab:mhsa-ablation} we show that both GAUs and Transformers are locally optimal on their own.

\section{Fast Linear Attention with GAU}
\label{sec:approx-attn}
There are two observations from Section~\ref{sec:quad} that motivate us to extend GAU to modeling long sequences:
\begin{itemize}[leftmargin=*,itemsep=0pt,topsep=0pt]
\item First, the gating mechanism in GAU allows the use of a weaker (single-headed, softmax-free) attention without quality loss.
If we further adapt this intuition into modeling long sequences with attention, GAU could also boost the effectiveness of approximate (weak) attention mechanisms such as local, sparse and linearized attention.
\item In addition, the number of attention modules is naturally doubled with GAU --- recall MLP+MHSA$\approx$2$\times$GAU in terms of cost (Section~\ref{sec:quad}).
Since approximate attention usually requires more layers to capture full dependency~\cite{dai2019transformer,child2019generating}, this property also makes GAU more appealing in handling long sequences.
\end{itemize}
With this intuition in mind, we start by reviewing some related work on modeling long sequences with attention, and then show how we enable GAU to achieve Transformer-level quality in linear time on long sequences.

\subsection{Existing Linear-Complexity Variants}
\paragraph{Partial Attention.} A popular class of methods tries to approximate the full attention matrix with different partial/sparse patterns, including local window~\cite{dai2019transformer,rae2019compressive}, local+sparse~\cite{child2019generating,li2019enhancing,beltagy2020longformer,zaheer2020big}, axial~\cite{ho2019axial,huang2019ccnet}, learnable patterns through hashing~\cite{kitaev2020reformer} or clustering~\cite{roy2021efficient}.
Though not as effective as full attention, these variants are usually able to enjoy quality gains from scaling to longer sequences.
However, the key problem with this class of methods is that they involve extensive irregular or regular memory re-formatting operations such as gather, scatter, slice and concatenation, which are not friendly to modern accelerators of massive parallelism, particularly specialized ASICs like TPU.
As a result, their practical benefits (speed and RAM efficiency), if any, largely depend on the choice of accelerator and usually fall behind the theoretical analysis.
Hence, in this work, we deliberately minimize the number of memory re-formatting operations in our model.

\begin{figure}[t]
\centering
\includegraphics[width=0.98\linewidth]{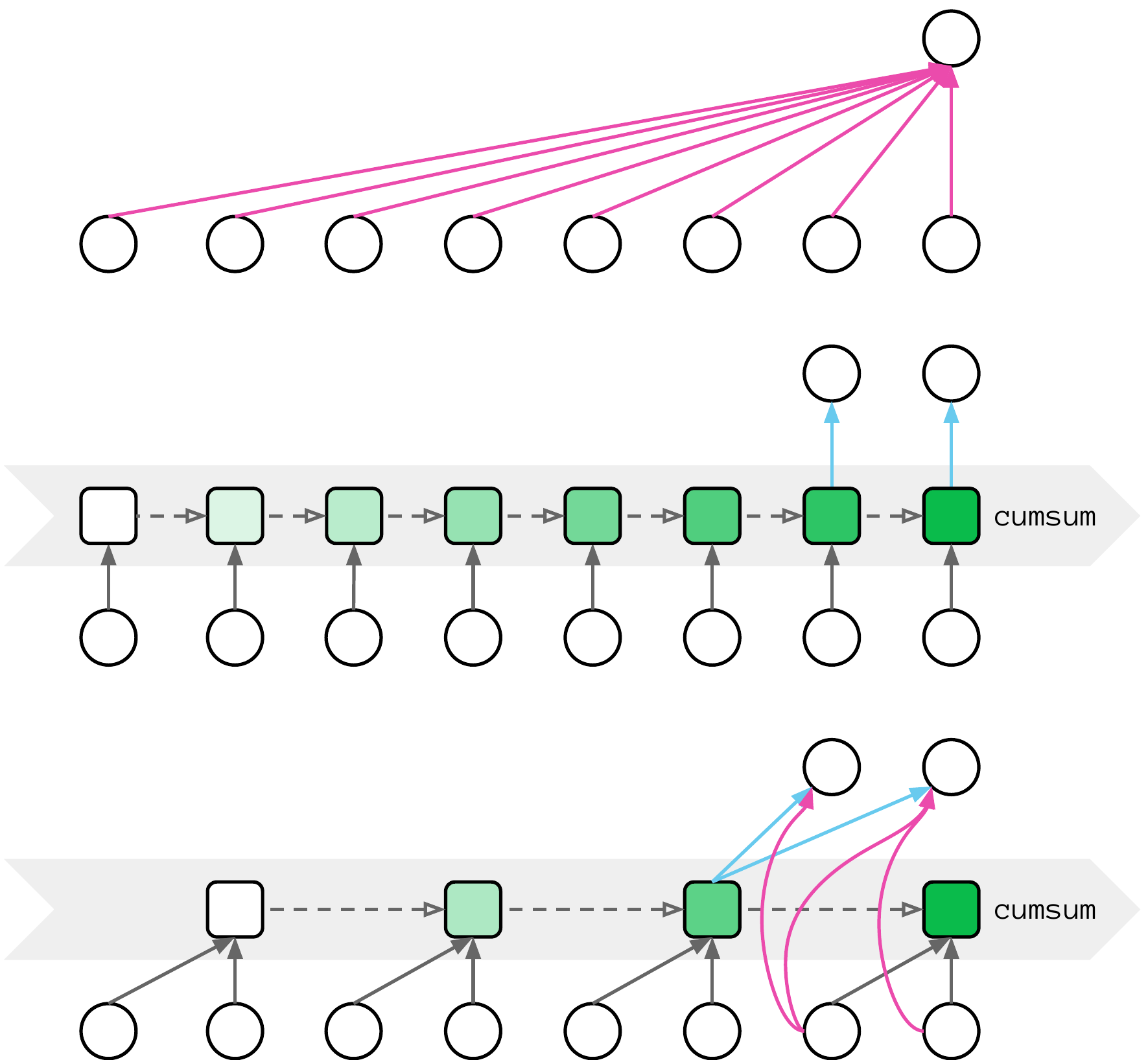}\vspace{0.5em}
\caption{(\textbf{top}) Quadratic attention, (\textbf{mid}) Linear attention, (\textbf{bottom}) Proposed \attn with a chunk size ($C$) of 2 ($C$ is always greater than or equal to 128 in our experiments). Our method significantly reduces the compute in quadratic attention (red links), while requiring substantially less RNN-style steps (green squares) in conventional linear attention.}
\label{fig:linear-attention}
\end{figure}

\paragraph{Linear Attention.} Alternatively, another popular line of research linearizes the attention computation by decomposing the attention matrix and then re-arranging the order of matrix multiplications~\cite{choromanski2020rethinking,wang2020linformer,katharopoulos2020transformers,peng2021rfa}.
Schematically, the linear attention can be expressed as
\begin{align*}
    \hat{V}_\text{lin} = Q \underbrace{\left(K^\top V\right)}_{\mathbb{R}^{d \times d}} \ \xrightarrow{\text{approx}}\ \hat{V}_\text{quad} = \mathrm{Softmax}\underbrace{\left( Q K^\top \right)}_{\mathbb{R}^{T \times T}} V
\end{align*}
where $Q, K, V \in \mathbb{R}^{T \times d}$ are the query, key and value representations, respectively.
Re-arranging the computation reduces the complexity w.r.t $T$ from quadratic to linear.

Another desirable property of linear attention is its \textit{constant}\footnote{Constant is with respective to the sequence length $T$.} computation and memory for each \textit{auto-regressive decoding} step at inference time.
To see that, define $M_t = K_{:t}^\top V_{:t}$ and notice that the computation of $M_t$ can be fully \textit{incremental}:
\begin{align}
\label{eq:incremental}
    M_t = M_{t-1} + K_t V_t^\top
\end{align}
This means we only need to maintain a cache with constant $\mathcal{O}(d^2)$ memory and whenever a new input arrives at time stamp $t$, only constant $\mathcal{O}(d^2)$ computation is required to accumulate $K_t V_t^\top$ into $M_{t-1}$ and get $M_t$.
On the contrary, full quadratic attention requires linear $\mathcal{O}(Td)$ computation and memory for each decoding step, as each new input has to attend to all the previous steps.

However, on the other hand, re-arranging the computation in linear attention leads to a severe inefficiency during auto-regressive training.
As shown in Fig.~\ref{fig:linear-attention} (mid), due to the causal constraint for auto-regressive training, the query vector at each time step $Q_t$ corresponds to a different cache value $M_t = K_{:t}^\top V_{:t}$.
This requires the model to compute and cache $T$ different values $\{M_t\}_{t=1}^{T}$ instead of only one value $K^\top V$ in the non-autoregressive case.
\textit{In theory}, the sequence $\{M_t\}_{t=1}^{T}$ can be obtained in $\mathcal{O}(Td^2)$ by first computing $\{K_t V_t^\top\}_{t=1}^{T}$ and then performing a large cumulative sum (\texttt{cumsum}) over $T$ tokens.
But in practice, the \texttt{cumsum} introduces an RNN-style \textit{sequential dependency} of $T$ steps,
where an $\mathcal{O}(d^2)$ state needs to be processed each step.
The sequential dependency not only limits the degree of parallelism, but more importantly requires $T$ \textit{memory access} in the loop, which usually costs much more time than computing the element-wise addition on modern accelerators.
As a result, there exists a considerable gap between the theoretical complexity and actual running time.
In practice, we find that directly computing the full quadratic attention matrix is even faster than the re-arranged (linearized) version on both TPUs (Figure~\ref{fig:lm_wiki_lat}) and GPUs (Appendix~\ref{appendix:gpu_training}).

\subsection{Our Method: Mixed Chunk Attention}
Based on the strengths and weaknesses of existing linear-complexity attentions, we propose \textit{\attn}, which merges the benefits from both partial attention and linear attention.
The high-level idea is illustrated in Figure~\ref{fig:linear-attention}.
Below we reformulate GAU to incorporate this idea.

\paragraph{Preparation.} The input sequence is first chunked into $G$ \textit{non-overlapping} chunks of size $C$, i.e. $[T] \to [T/C \times C]$.
Then, $U_g \in \mathbb{R}^{C \times e}$, $V_g \in \mathbb{R}^{C \times e}$ and $Z_g \in \mathbb{R}^{C \times s}$ are produced for each chunk $g$ following the GAU formulation in Eq. \eqref{eq:glu-uv} and Eq.
\eqref{eq:gau-key}.
Next,
four types of attention heads $Q_g^\text{quad}$, $K_g^\text{quad}$, $Q_g^\text{lin}$, $K_g^\text{lin}$ are produced from $Z_g$ by applying per-dim scaling and offset (this is very cheap).

We will describe how GAU's attention can be efficiently approximated using a local attention plus a global attention.
Note all the major tensors $U_g$, $V_g$ and $Z_g$ are shared between the two components.
The only additional parameters introduced over the original GAU are the per-dim scalars and offsets for generating $Q_g^\text{lin}$ and $K_g^\text{lin}$ (4$\times s$ parameters).

\paragraph{Local Attention per Chunk.}
First, a local quadratic attention is independently applied to each chunk of length $C$ to produce part of the pre-gating state:
\begin{align*}
    \hat{V}_{g}^\text{quad} = \mathrm{relu}^2\Big( Q_g^\text{quad} {K_g^\text{quad}}^\top + b \Big) V_{g}.
\end{align*}
The complexity of this part is $\mathcal{O}(G \times C^2\times d) = \mathcal{O}(TCd)$, which is linear in $T$ given that $C$ remains constant.

\paragraph{Global Attention across Chunks.}
In addition, a global linear attention mechanism is employed to capture long-range interaction across chunks
\begin{align}
    \label{eq:non-causal}
    \text{Non-Causal:} && \hat{V}_g^\text{lin} &= Q_g^\text{lin} \bigg( \sum_{h=1}^G {K_h^\text{lin}}^\top V_h \bigg), \\
    \label{eq:causal}
    \text{Causal:} && \hat{V}_g^\text{lin} &= Q_g^\text{lin} \bigg( \sum_{h=1}^{g-1} {K_h^\text{lin}}^\top V_h \bigg).
\end{align}
Note the summations in Eq.~\eqref{eq:non-causal} and Eq.~\eqref{eq:causal} are performed at the \emph{chunk} level.
For the causal (auto-regressive) case,
this reduces the number of elements in the \texttt{cumsum} in token-level linear attention by a factor of $C$ (a typical $C$ is 256 in our experiments),
leading to a significant training speedup.

Finally, $\hat{V}_g^\text{quad}$ and $\hat{V}_g^\text{lin}$ are added together, followed by gating and a post-attention projection analogous to Eq.~\eqref{eq:gau-gating}:
\begin{align*}
    O_g = \left[ U_g \odot  \left(\hat{V}_{g}^\text{quad} + \hat{V}_{g}^\text{lin} \right) \right] W_o.
\end{align*}

The \attn is simple to implement and the corresponding pseudocode is given in Code~\ref{code:proposed}.
\begin{figure}
\begin{lstlisting}[language=python, mathescape]
def _global_linear_attn(q, k, v, causal):
    if causal:
        kv = tf.einsum('bgcs,bgce$\rightarrow$bgse', k, v)
        kv = tf.cumsum(kv, axis=1, exclusive=True)
        return tf.einsum('bgcs,bgse$\rightarrow$bgce', q, kv)
    else:
        kv = tf.einsum('bgcs,bgce$\rightarrow$bse', k, v)
        return tf.einsum('bgcs,bse$\rightarrow$bgce', q, kv)

def _local_quadratic_attn(q, k, v, causal):
    qk = tf.einsum('bgns,bgms$\rightarrow$bgnm', q, k)
    a = relu(qk + rel_pos_bias(q, k)) ** 2
    a = causal_mask(a) if causal else a
    return tf.einsum('bgnm,bgme$\rightarrow$bgne', a, v)

def attn(x, v, causal, s=128):
    # x: [B x G x C x D]; v: [B x G x C x E]
    z = dense(x, s)
    v_quad = _local_quadratic_attn(
        scale_offset(z), scale_offset(z), v, causal)
    v_lin = _global_linear_attn(
        scale_offset(z), scale_offset(z), v, causal)
    return v_quad + v_lin
\end{lstlisting}
\captionof{Code}{Pseudocode for \attn.}
\label{code:proposed}
\end{figure}

\subsubsection{Discussions}
\paragraph{Fast Auto-regressive Training.} Importantly, as depicted in Fig.~\ref{fig:linear-attention} (bottom), thanks to chunking, the sequential dependency in the auto-regressive case reduces from $T$ steps in the standard linear attention to $G = T / C$ steps in the chunked version in Eq.~\eqref{eq:causal}.
Therefore, we observe the auto-regressive training becomes dramatically faster with the chunk size is in $\{128, 256, 512\}$.
With the inefficiency of auto-regressive training eliminated, the proposed model still enjoys the constant per-step decoding memory and computation of $\mathcal{O}(Cd^2)$, where the additional constant $C$ comes from the local quadratic attention.

\begin{figure*}[!ht]
\centering     %
\vspace{-0.5cm}
\resizebox{1.0\textwidth}{!}{
\subfigure
{
    \label{fig:mlm_legend}
    \includegraphics[ width=0.35\textwidth]{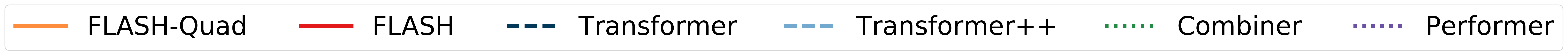}
}
}
\vspace{-0.8cm}
\\
\setcounter{subfigure}{0}
\resizebox{\textwidth}{!}{
\subfigure[Per-step training latency]
{
    \label{fig:mlm_c4_lat}
    \includegraphics[height=4 cm, width=0.31\textwidth]{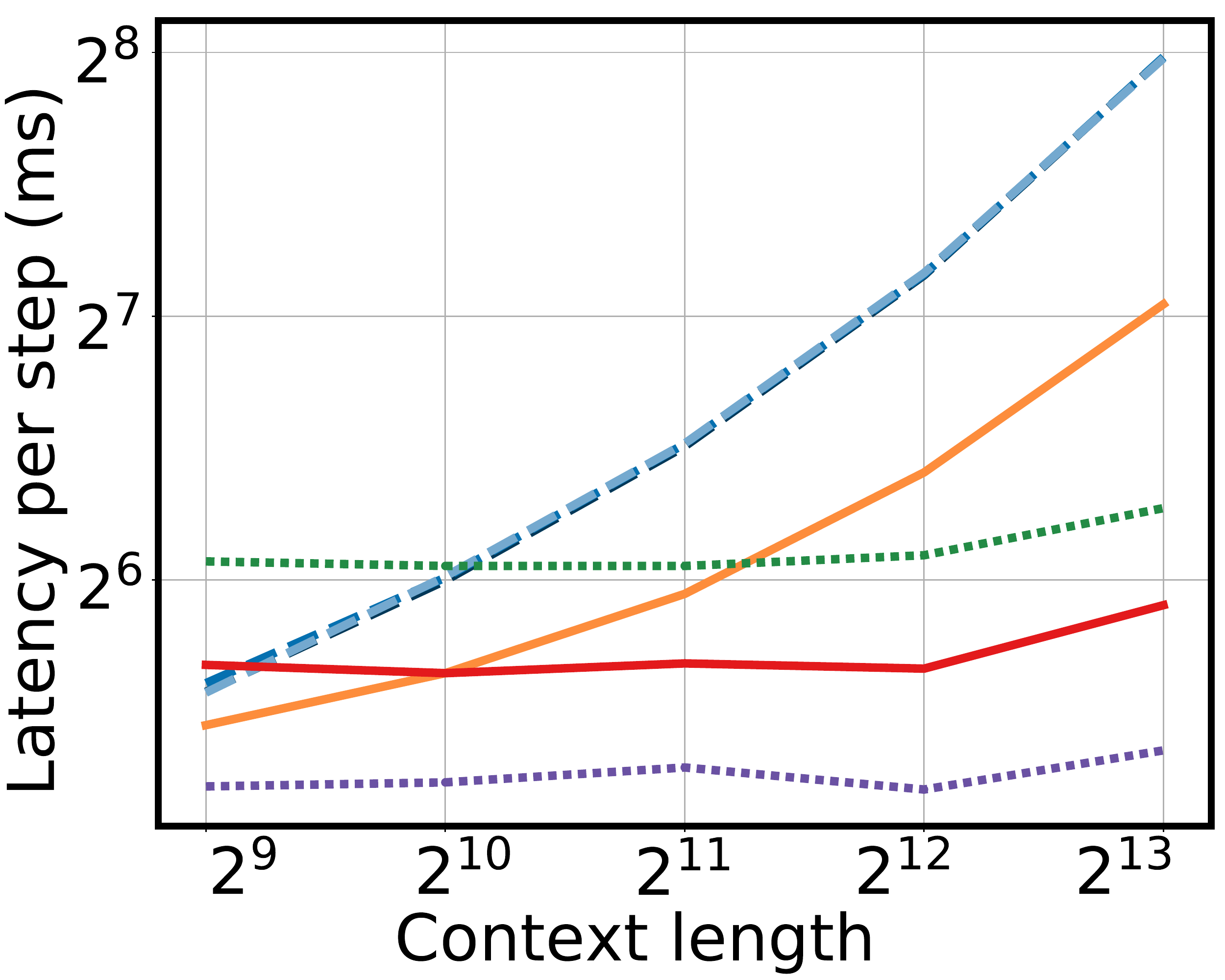}
}%
\subfigure[Context length = 512]
{
    \label{fig:mlm_c4_s512}
    \includegraphics[height=4.22 cm, width=0.33\textwidth]{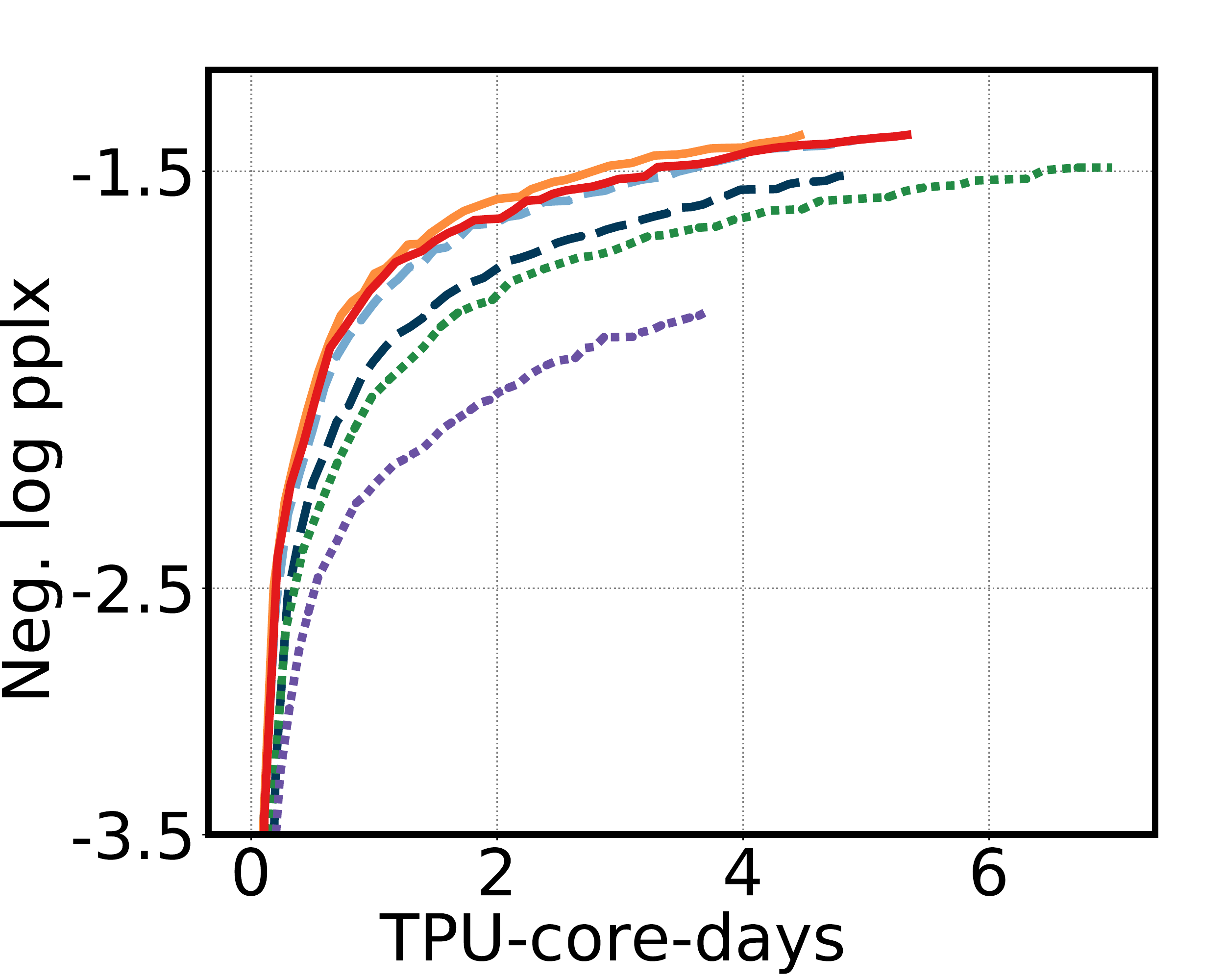}
}%
\subfigure[Context length = 1024]
{
    \label{fig:mlm_c4_s1024}
    \includegraphics[height=4.22 cm, width=0.33\textwidth]{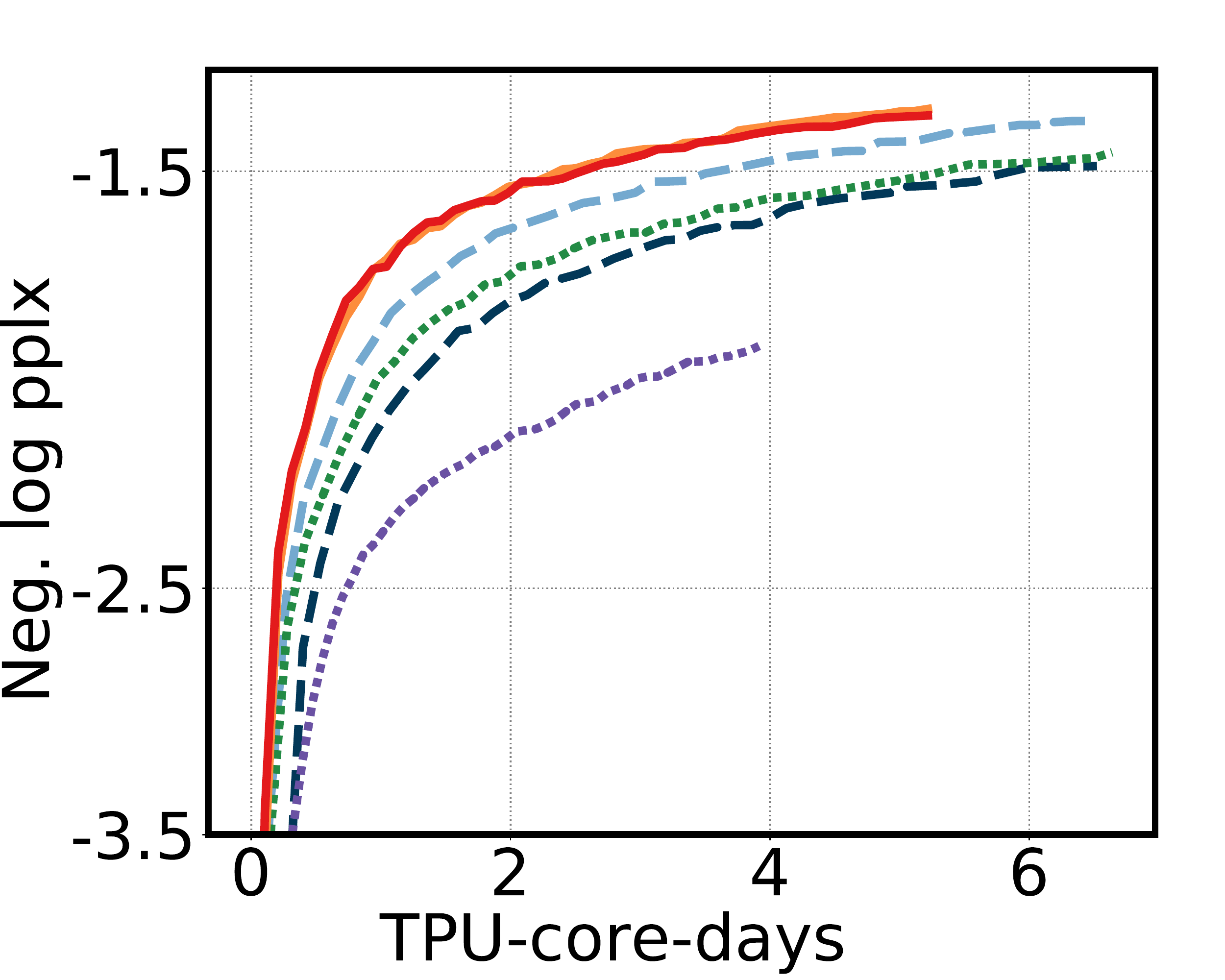}
}
}
\\
\resizebox{\textwidth}{!}{
\subfigure[Context length = 2048]
{
    \label{fig:mlm_c4_s2048}
    \includegraphics[height=4.22 cm, width=0.33\textwidth]{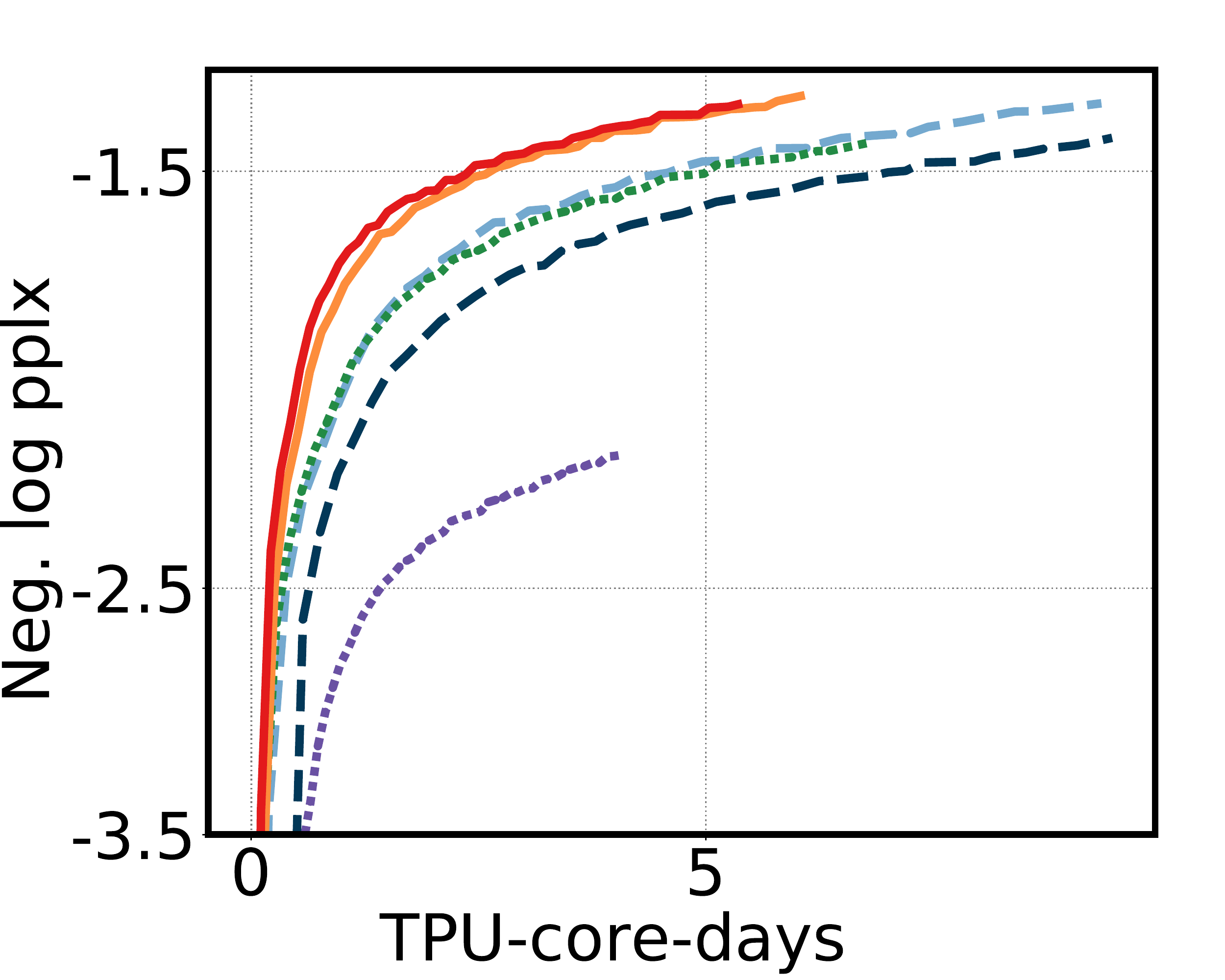}
}%
\subfigure[Context length = 4096]
{
    \label{fig:mlm_c4_s4096}
    \includegraphics[height=4.22 cm, width=0.33\textwidth]{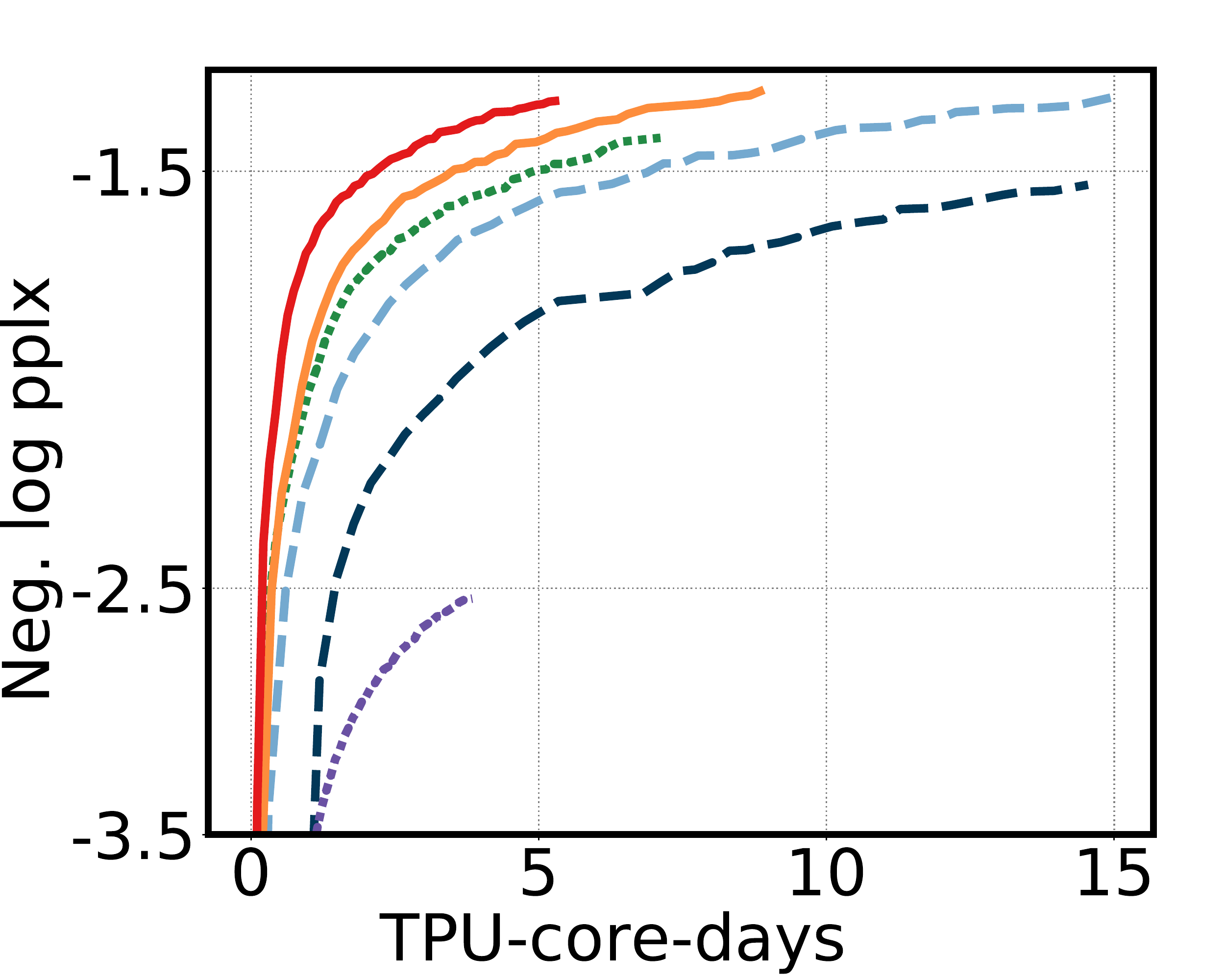}
}%
\subfigure[Context length = 8192]
{
    \label{fig:mlm_c4_s8192}
    \includegraphics[height=4.22 cm, width=0.33\textwidth]{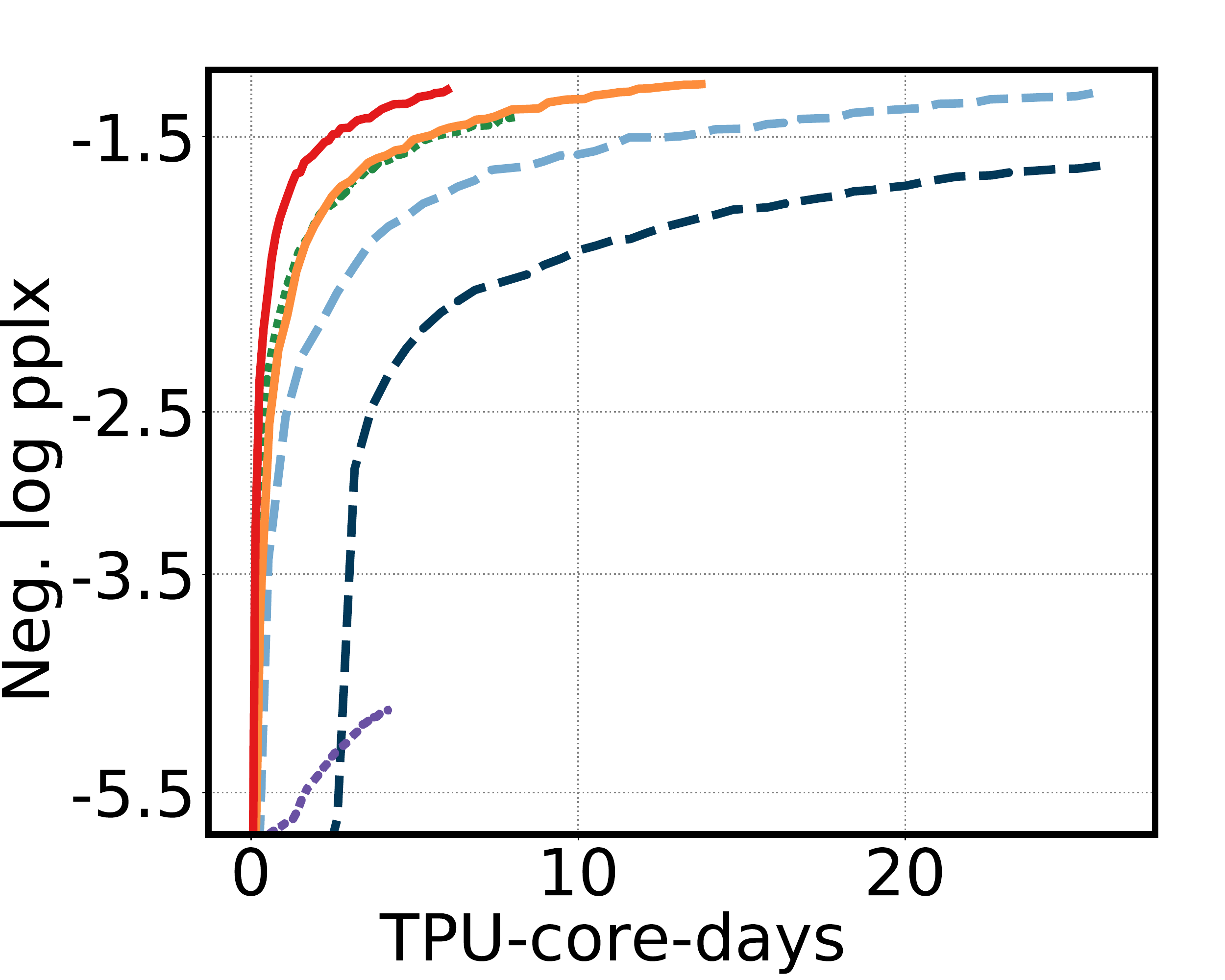}
}
}
\caption{Masked language modeling validation-set results on the C4 dataset --- \small{All models are comparable in size at around 110M (i.e., BERT-Base scale) and trained for 125K steps with 2$^{18}$ tokens per batch. The quality is measured in negative log perplexity.}}
\label{fig:mlm}
\end{figure*}

\paragraph{On Non-overlapping Local Attention.} Chunks in our method does not overlap with each other. In theory, instead of using the non-overlapping local attention, any partial attention variant could be used as a substitute while keeping the chunked linear attention fixed.
As a concrete example, we explored allowing each chunk to additionally attends to its nearby chunks, which essentially makes the local attention overlapping, similar to Longformer~\cite{beltagy2020longformer} and BigBird~\cite{zaheer2020big}.
While overlapping local attention consistently improves quality, it also introduces many memory re-formatting operations that clearly harm the actual running speed. 
In our preliminary experiments with language modeling on TPU, we found the cost-benefit trade-off of using overlapping local attention may not be as good as adding more layers in terms of both memory and speed.
In general, we believe the optimal partial attention variant is task-specific, while non-overlapping local attention is always a strong candidate when combined with the choice of chunked linear attention.

\paragraph{Connections to Combiner.} Similar to our method, Combiner~\cite{ren2021combiner} also splits the sequence into non-overlapping chunks and utilizes quadratic local attention within each chunk. The key difference lies in how the long-range information is summarized and combined with the local information (e.g., our \attn allows larger effective memory per chunk hence leads to better quality).
See Appendix~\ref{sec:combiner} for detailed discussions.

\section{Experiments}
\label{sec:experiment}
We focus on two of our models that have different complexities with respect to the context length.
The quadratic-complexity model \qname refers to a stack of GAUs whereas the linear-complexity model named \lname consists of both GAUs and the proposed \attn.
To demonstrate their efficacy and general applicability, we evaluate them on both bidirectional and auto-regressive sequence modeling tasks over multiple large-scale datasets.

\paragraph{Baselines.} First of all, the vanilla Transformer~\cite{vaswani2017attention} with GELU activation function~\cite{hendrycks2016gelu} is included as a standard baseline for calibration.
Despite of being a popular baseline in the literature,
we find that RoPE~\cite{su2021rope} and GLU~\cite{shazeer2020glu} can lead to significant performance boosts.
We therefore also include Transformer + RoPE (Transformer+) and Transformer + RoPE + GLU (Transformer++) as two much stronger baselines with quadratic complexity.

\begin{figure*}[ht]
\centering     %
\vspace{-0.5cm}
\resizebox{1.0\textwidth}{!}{
\subfigure
{
    \label{fig:lm_legend}
    \includegraphics[ width=0.35\textwidth]{figures/legend.pdf}
}
}
\vspace{-0.8cm}
\\
\setcounter{subfigure}{0}
\resizebox{\textwidth}{!}{
\subfigure[Per-step training latency]
{
    \label{fig:lm_wiki_lat}
    \includegraphics[width=0.32\textwidth]{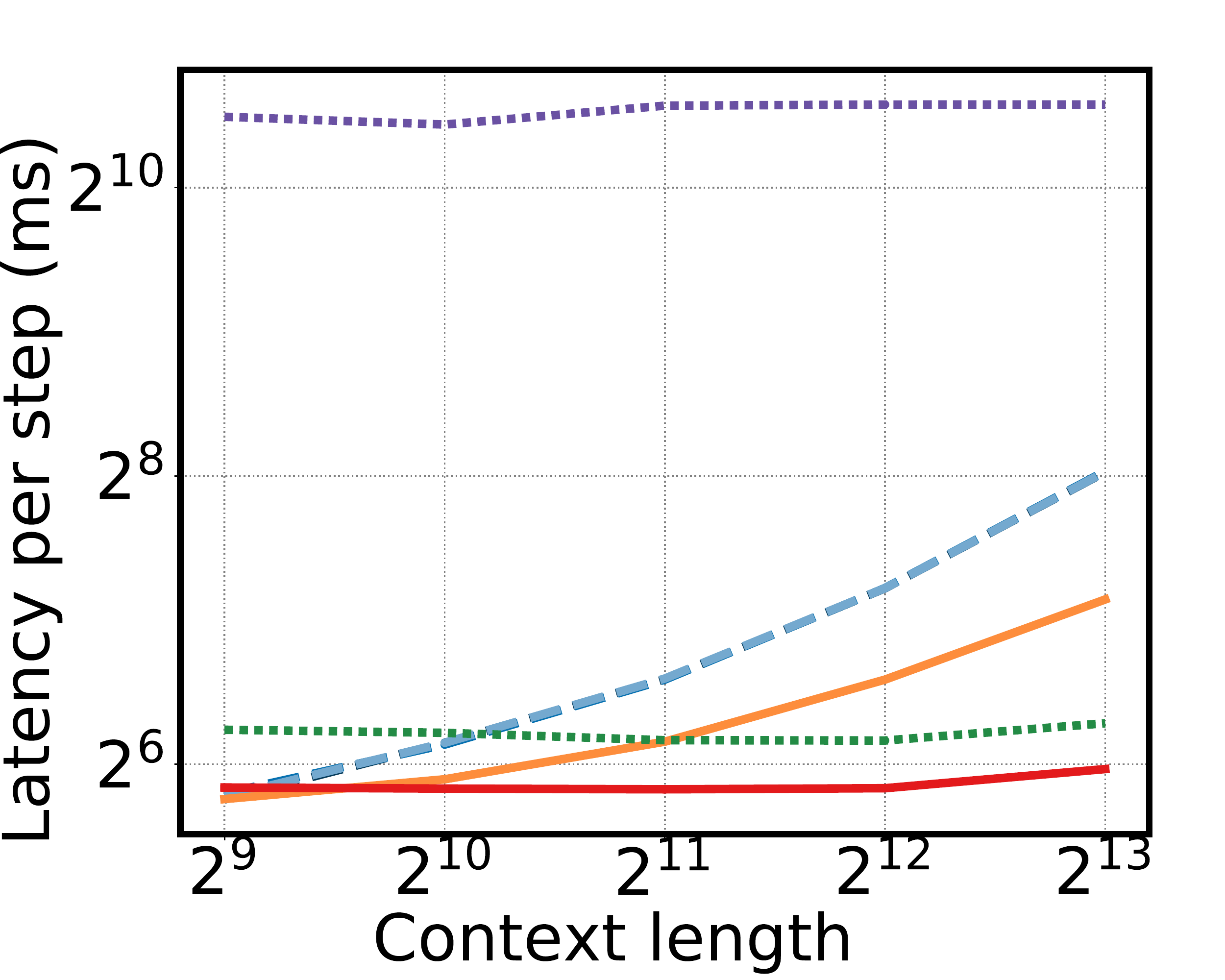}
}%
\subfigure[Context length = 512]
{
    \label{fig:lm_wiki_s512}
    \includegraphics[width=0.33\textwidth]{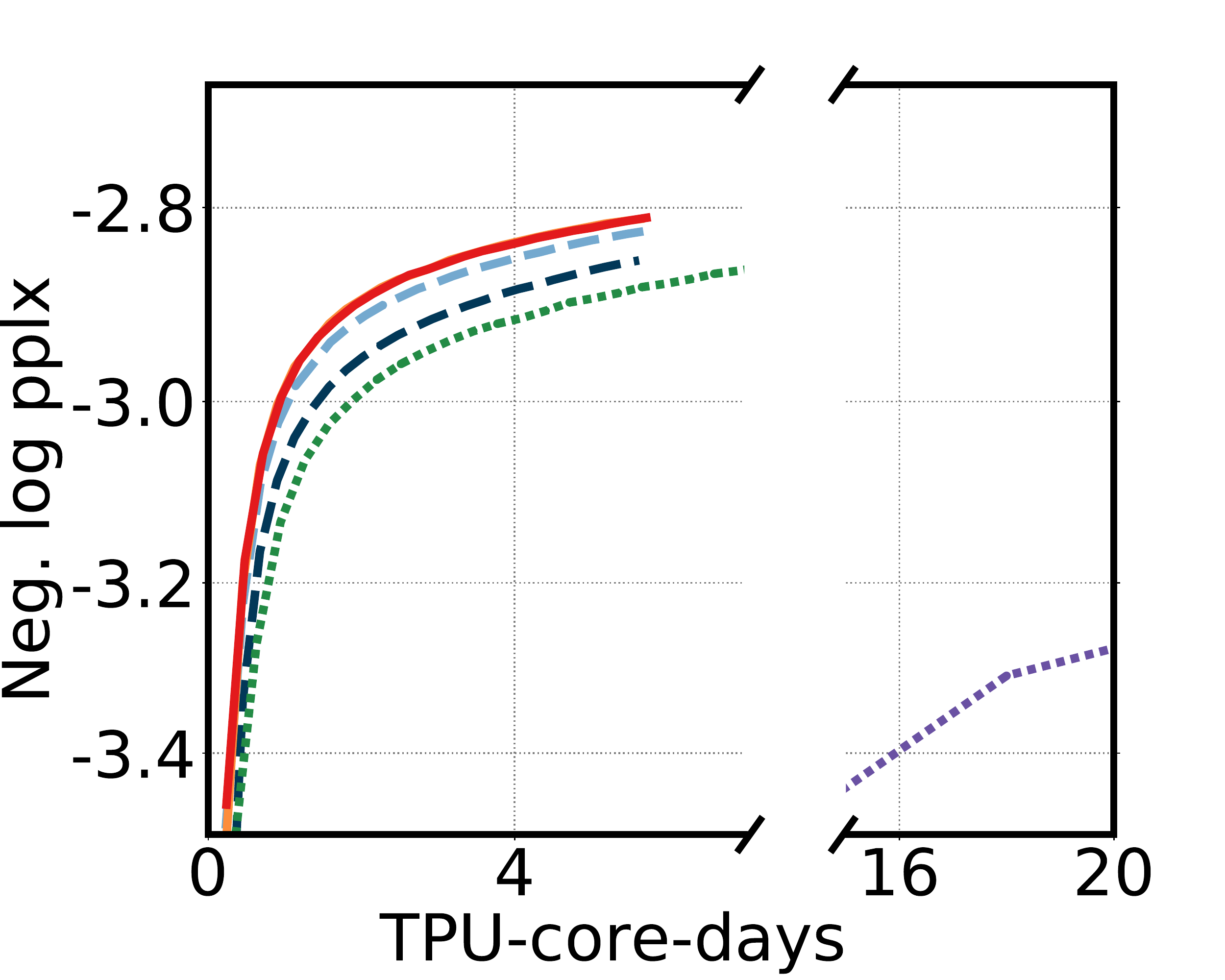}
}%
\subfigure[Context length = 1024]
{
    \label{fig:lm_wiki_s1024}
    \includegraphics[ width=0.33\textwidth]{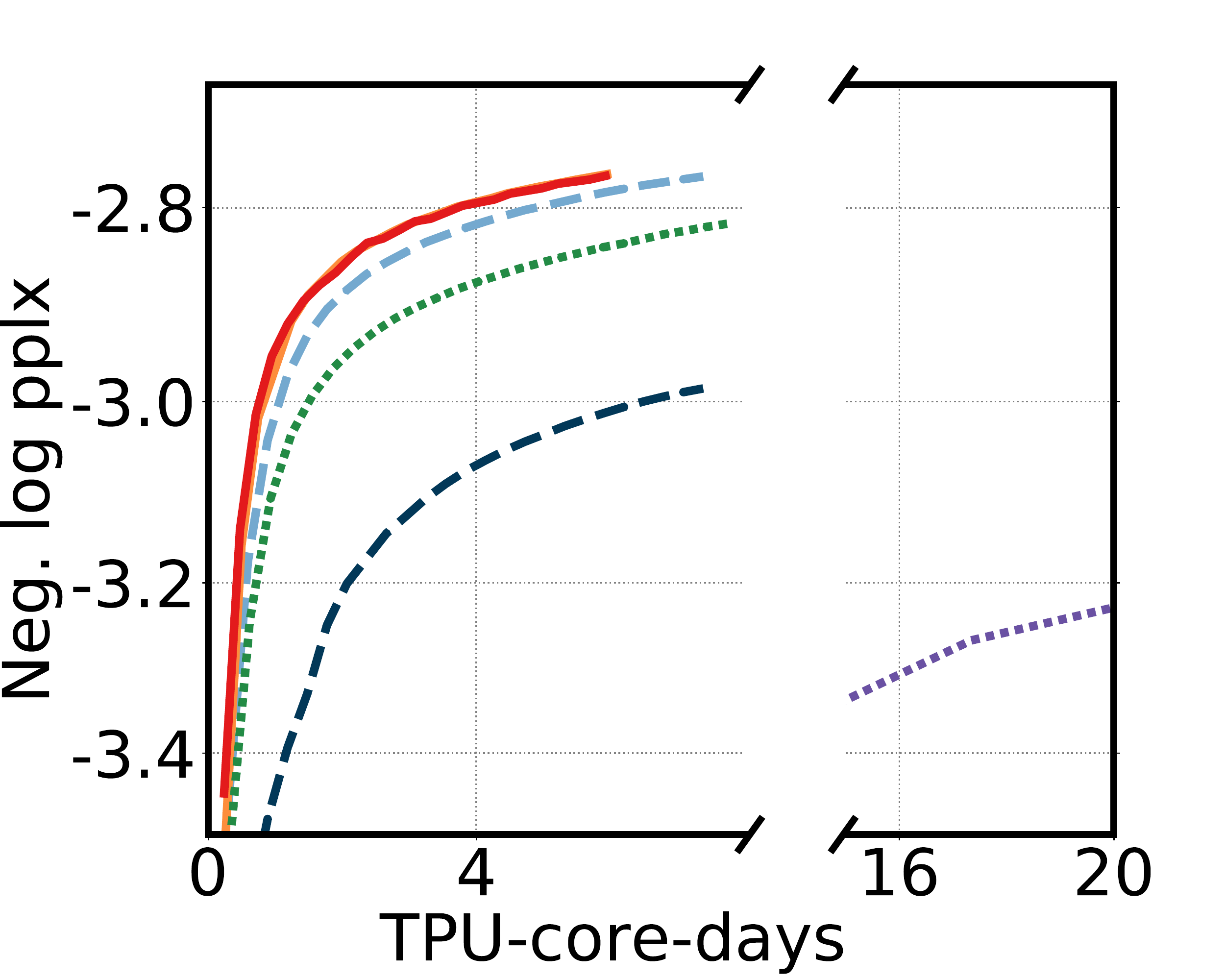}
}
}
\\
\resizebox{\textwidth}{!}{
\subfigure[Context length = 2048]
{
    \label{fig:lm_wiki_s2048}
    \includegraphics[width=0.33\textwidth]{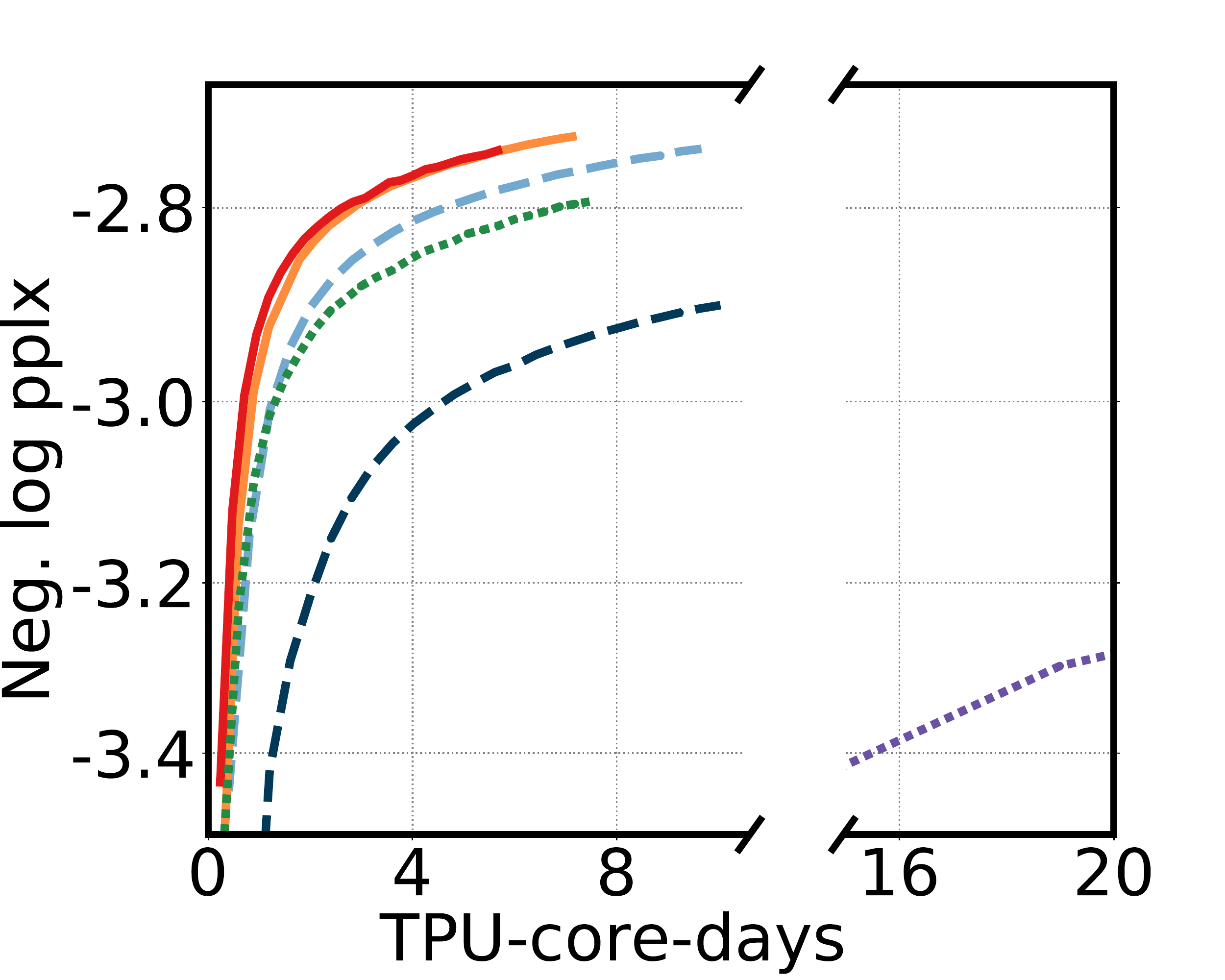}
}%
\subfigure[Context length = 4096]
{
    \label{fig:lm_wiki_s4096}
    \includegraphics[width=0.33\textwidth]{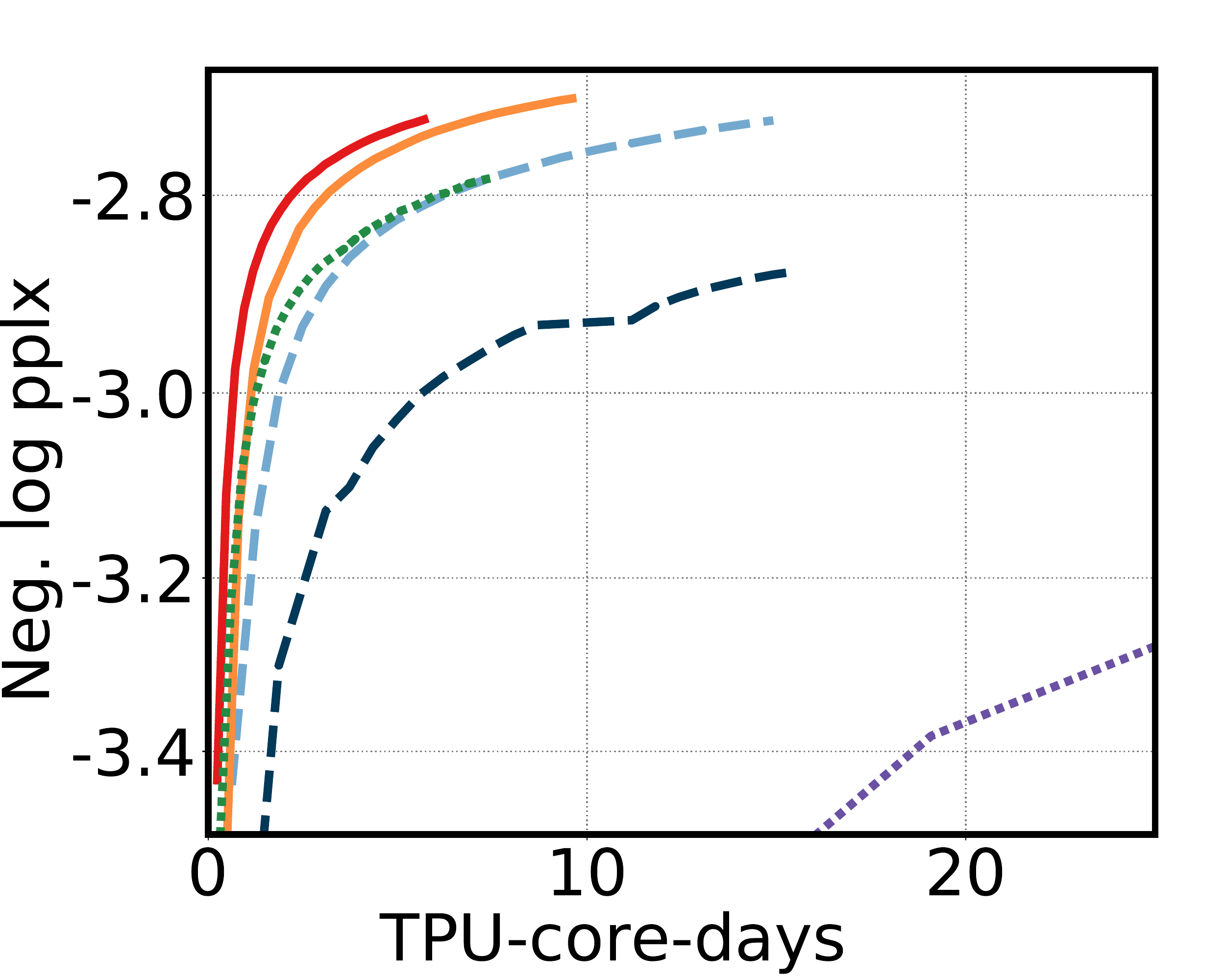}
}%
\subfigure[Context length = 8192]
{
    \label{fig:lm_wiki_s8192}
    \includegraphics[ width=0.33\textwidth]{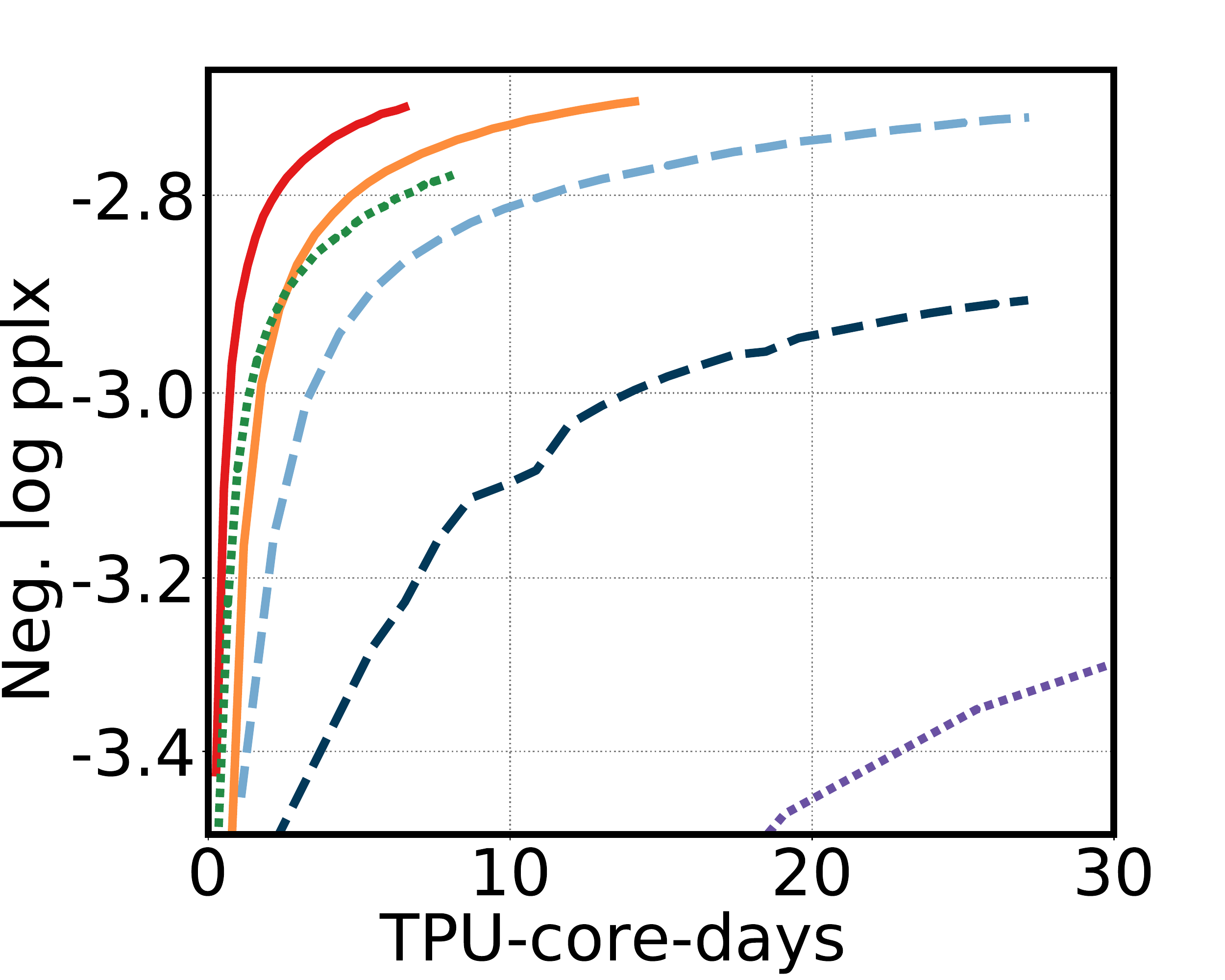}
}
}
\caption{Auto-regressive language modeling validation-set results on the Wiki-40B dataset --- \small{All models are sized around 110M (i.e., BERT-Base scale) and trained for 125K steps with 2$^{18}$ tokens per batch. The quality is measured in negative log perplexity.}}
\label{fig:lm}
\end{figure*}

To demonstrate the advantages of our models on long sequences, we further compare our models with two notable linear-complexity Transformer variants---Performer~\cite{choromanski2020rethinking} and Combiner~\cite{ren2021combiner}, where Performer is a representative linear attention method and Combiner (using a chunked attention design similar to ours) has shown superior cost-benefit trade-off over many other approaches~\cite{ren2021combiner}.
To get the best performance, we use the rowmajor-axial variant of Combiner (Combiner-Axial) and the ReLU-kernel variant of Performer.
Both models are also augmented with RoPE.

For fair comparison, all models are implemented in the same codebase to ensure identical tokenizer and hyper-parameters for training and evaluation.
The per-step training latencies of all models are measured using TensorFlow Profiler.
See Appendix \ref{appendix:setup} for detailed settings and model specifications.

\subsection{Bidirectional Language Modeling}
In BERT~\cite{devlin2018bert}, masked language modeling (MLM) reconstructs randomly masked out tokens in the input sequence.
We pretrain and evaluate all models on the C4 dataset~\cite{raffel2020c4}.
We consistently train each model with $2^{18}$ tokens per batch for 125K steps, while varying the context length on a wide range including 512, 1024, 2048, 4096, and 8192.
The quality of each model is reported in perplexity as a proxy metric for the performance on downstream tasks.
The training speed of each model (i.e., training latency per step) is measured with 64 TPU-v4 cores, and the total training cost is reported in TPU-v4-core-days.

Figure~\ref{fig:mlm_c4_lat} shows the latency of each training step for all models at different context lengths.
Results for Transformer+ are omitted for brevity as it lies in between Transformer and Transformer++.
Across all the six models, latencies for Combiner, Performer, and~\lname remain roughly constant as the context length increases, demonstrating linear complexity with respect to context length.
\qname is consistently faster than Transformer and Transformer++ for all context lengths.
In particular, \qname is 2$\times$ as fast as Transformer++ when the context length increases to 8192.
More importantly, as shown in Figures~\ref{fig:mlm_c4_s512}-\ref{fig:mlm_c4_s8192}, for all sequence lengths ranging from 512 to 8192, our models always achieve the best quality (i.e., lowest perplexity) under the same computational resource.
In particular, if the goal is to match Transformer++'s final perplexity at step 125K, \qname and \lname can reduce the training cost by 1.1$\times$--2.5$\times$ and 1.0$\times$--4.8$\times$, respectively.
It is worth noting that, to the best of our knowledge, \lname is the only linear-complexity model that achieves perplexity competitive with the fully-augmented Transformers and its quadratic-complexity counterpart.
See Appendix~\ref{appendix:tabular_results} for a detailed quality and speed comparison of all models.

\subsection{Auto-regressive Language Modeling}
\begin{table*}[ht]
\caption{Auto-regressive language models on the PG-19 dataset --- \small{Latency (Lat.) is measured with 64 TPU-v4 cores.}}
\label{tbl:pg19}
\begin{threeparttable}
\resizebox{1.0\textwidth}{!}{
\begin{tabular}{@{}l|cccccccccccc@{}}
\toprule
\multirow{3}{*}{Model} & \multicolumn{12}{c}{Context Length} \\ \cmidrule(l){2-13} 
 & \multicolumn{3}{c|}{1024} & \multicolumn{3}{c|}{2048} & \multicolumn{3}{c|}{4096} & \multicolumn{3}{c}{8192} \\ \cmidrule(l){2-13} 
 & PPLX & Lat. & \multicolumn{1}{c|}{Speedup*} & PPLX & Lat. & \multicolumn{1}{c|}{Speedup*} & PPLX & Lat. & \multicolumn{1}{c|}{Speedup*} & PPLX & Lat. & Speedup* \\ \midrule
\multicolumn{1}{l|}{Transformer+} & 44.45 & 282 & \multicolumn{1}{c|}{1.00$\times$} & 43.14 & 433 & \multicolumn{1}{c|}{1.00$\times$} & 42.80 & 698 & \multicolumn{1}{c|}{1.00$\times$} & 43.27 & 1292 & 1.00$\times$ \\
\multicolumn{1}{l|}{Transformer++} & 44.47 & 292 &  \multicolumn{1}{c|}{--} & 43.18 & 441 & \multicolumn{1}{c|}{--} & 43.13 & 712 & \multicolumn{1}{c|}{--} & 43.26 & 1272 & 1.21$\times$ \\
\multicolumn{1}{l|}{Combiner} & 46.04 & 386 & \multicolumn{1}{c|}{--} & 44.68 & 376 & \multicolumn{1}{c|}{--} & 43.99 & 374 & \multicolumn{1}{c|}{--} & 44.12 & 407 & -- \\ \midrule
\multicolumn{1}{l|}{\qname} & \textbf{43.40} & \textbf{231} & \multicolumn{1}{c|}{\textbf{2.18$\times$}} & \textbf{42.01} & 273 & \multicolumn{1}{c|}{$3.29\times$} & 41.46 & 371 & \multicolumn{1}{c|}{$3.59\times$} & 41.68 & 560 & $5.23\times$ \\
\multicolumn{1}{l|}{\lname} & 44.06 & \textbf{234} & \multicolumn{1}{c|}{\textbf{$1.66\times$}} & 42.17 & \textbf{237} & \multicolumn{1}{c|}{\textbf{3.85$\times$}} & \textbf{40.72} & \textbf{234} & \multicolumn{1}{c|}{\textbf{6.75$\times$}} & \textbf{41.07} & \textbf{250} & \textbf{12.12$\times$} \\ \bottomrule
\end{tabular}
}
\begin{tablenotes}
  \item[*] \small{Measured based on time taken to match Transformer+'s final quality (at step 125K) on TPU.} 
  \item[--] \small{Indicates that the specific model fails to achieve the same perplexity as Transformer+.}
\end{tablenotes}
\end{threeparttable}
\end{table*}

For auto-regressive language modeling, we focus on the Wiki-40B~\cite{guo2020wiki} and PG-19~\cite{rae2019compressive} datasets, which consist of clean English Wikipedia pages and books extracted from Project Gutenberg, respectively.
It is worth noting that the average document length in PG-19 is 69K words, making it ideal for evaluating model performance over long context lengths.
We train and evaluate all models with $2^{18}$ tokens per batch for 125K steps, with context lengths ranging from 512 to 8K for Wiki-40B and 1K to 8K for PG-19.
We report token-level perplexity for Wiki-40B and word-level perplexity for PG-19.

Figure~\ref{fig:lm_wiki_lat} shows that \qname and \lname achieve the lowest latency among quadratic and linear complexity models, respectively.
We compare the quality and training cost trade-offs of all models on Wiki40-B over increasing context lengths in Figures~\ref{fig:lm_wiki_s512}-\ref{fig:lm_wiki_s8192}.
Similar to the findings on MLM tasks, our models dominate all other models in terms of quality-training speed for all sequence lengths.
Specifically, \qname reduces the training time of Transformer++ by 1.2$\times$ to 2.5$\times$ and \lname cuts the compute cost by 1.2$\times$ to 4.9$\times$ while reaching a similar perplexity as Transformer++. Between our own models, \lname closely tracks the perplexity of \qname and starts to achieve a better perplexity-cost trade-off when the context length goes beyond 2048.
Detailed quality and speed comparisons for all models are included in Appendix~\ref{appendix:tabular_results}.

For PG-19, following \citeauthor{rae2019compressive}, an increased model scale of roughly 500M parameters (see Table \ref{tab:model_spec_pg19}) is used for all models in comparison.
The results are summarized in Table~\ref{tbl:pg19}.
Compared to the numbers in Wiki-40B, \name achieves a more pronounced improvements in perplexity and training time over the augmented Transformers on PG-19.
For example, with a context length of 8K, \qname and \lname are able to reach the final perplexity (at 125K-step) of Transformer+ in only 55K and 55K steps, yielding 5.23$\times$ and 12.12$\times$ of speedup, respectively.
We hypothesize that the increased gains over Transformer+ arise from the long-range nature of PG-19 (which consists of books).
Similar to our previous experiments, \lname achieves a lower perplexity than all of the full-attention Transformer variants while being significantly faster, demonstrating the effectiveness of our efficient attention design.

\subsection{Fine-tuning}
To demonstrate the effectiveness of FLASH over downstream tasks, we fine-tune our pre-trained models on the TriviaQA dataset~\cite{joshi2017triviaqa}. Passages in TriviaQA can span multiple documents, which challenges the capability of the models in handling long contexts.
For a fair and meaningful comparison, we pretrain all models on English Wikipedia (same domain as TriviaQA) with a context length of 4096 and a batch size of 64 for 125k steps.
For fine-tuning, we sweep over three different learning rates, including $1e^{-4}$, $7e^{-5}$, and $5e^{-5}$, and report the best validation-set F1 score across these runs.

\begin{table}[!ht]
\centering
\small
\caption{Results on TrivialQA with context length 4096 --- \small{``PT`` stands for pre-training and ``FT`` stands for fine-tuning. All models are comparable in size at around 110M. $s$ stands for the head size of the single-head attention. For FLASH, ``first-to-all'' means that we also let the first token in each chunk to attend to the entire sequence using a single-head softmax attention. Latency (Lat.) is measured with 32 TPU-v4 cores.}}
\begin{tabular}{@{}l|ccc@{}}
\toprule
\multirow{2}{*}{Model} & PT &  FT & PT / FT \\
 & PPLX & F1 & Lat. reduction \\ \midrule
Transformer+ & 3.48 & 74.2 & 1.00$\times$ / 1.00$\times$ \\
Combiner & 3.51 & 67.2 & \textbf{2.78$\times$ / 2.75$\times$} \\\midrule
FLASH-Quad\textsubscript{$s$=128} & 3.24 & 72.7 & 1.89$\times$ / 1.79$\times$ \\
FLASH-Quad\textsubscript{$s$=512} & \textbf{3.12} & \textbf{74.8} & 1.76$\times$ / 1.67$\times$ \\
FLASH\textsubscript{$s$=512} & 3.23 & 73.3 & 2.61$\times$ / 2.60$\times$ \\
FLASH\textsubscript{$s$=512} + first-to-all & 3.24 & 73.9 & \textbf{2.78$\times$ / 2.69$\times$} \\ \bottomrule
\end{tabular}
\label{tab:TrivaQA}
\end{table}

\begin{figure*}[ht]
\centering
\includegraphics[width=0.99\textwidth]{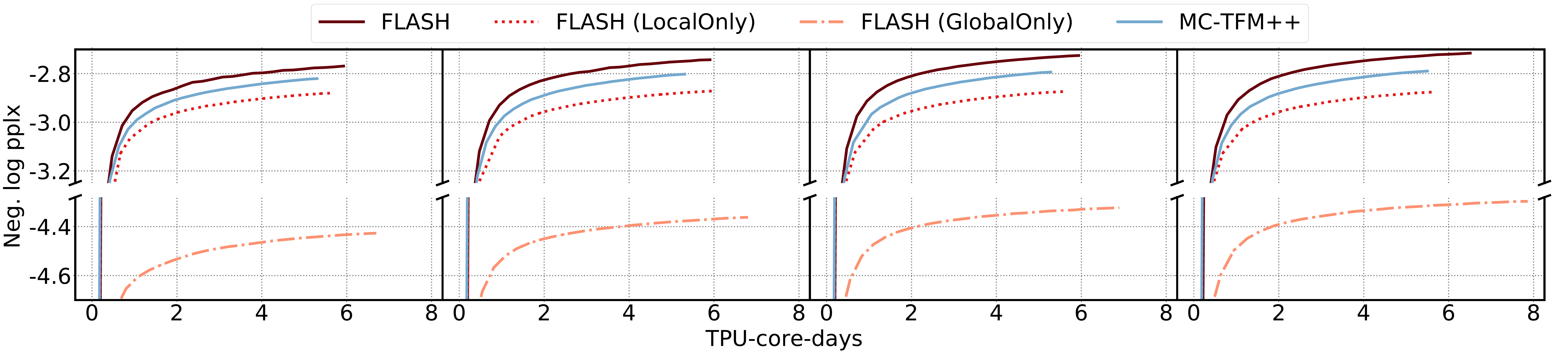}

\hspace*{0.3cm}\small{\hfil{(a) Context length = 1024}}
\small{\hfil{(b) Context length = 2048}}
\small{\hfil{(c) Context length = 4096}}
\small{\hfil{(d) Context length = 8192}}

\caption{Ablation study of the proposed \name architecture.
}
\label{fig:model_ablation}
\end{figure*}

We observe that the fine-tuning results of the FLASH family can benefit from several minor changes in the model configuration.
As shown in Table~\ref{tab:TrivaQA}, increasing the head size of FLASH-Quad from 128 to 512 leads to a significant boost of 2.1 point in the F1 score with negligible impact on speed.
We further identify several other tweaks that improve 
the linear FLASH variant specifically,
including using a small chunk size (128), disabling gradient clipping during finetuning, using softmax instead of squared ReLU for the [CLS] token, and (optionally) allowing the first token in each chunk to attend to the entire sequence using softmax.
With those changes, FLASH\textsubscript{$s$=512} achieves comparable quality to Transformer+ (0.3 difference in F1 is within the range of variance) while being 2.8$\times$ and 2.7$\times$ as fast as Transformer+ in pretraining and fine-tuning, respectively.

\subsection{Ablation Studies}
\paragraph{Significance of quadratic \& linear components.} To better understand the efficacy of \name, we first study how much the local quadratic attention and the global linear attention contribute to the performance individually.
To this end, we create \lname (LocalOnly) and \lname (GlobalOnly) by only keeping the local quadratic attention and the global linear attention in \lname, respectively.
In \lname (GlobalOnly), we reduce the chunk size from 256 to 64 to produce more local summaries for the global linear attention.
In Figure~\ref{fig:model_ablation} we see a significant gap between the full model and the two variants, suggesting that the linear and global attention are complementary to each other --- both are critical to the quality of the proposed \attn.

\paragraph{Significance of GAU.} Here we study the importance of using GAU in \lname.
To achieve this, \emph{we apply the same idea of \attn to Transformer++}. We refer to this variant as MC-TFM++ (MC stands for mixed chunk) which uses quadratic MHSA within each chunk and multi-head linear attention across chunks.
Effectively, MC-TFM++ has the same linear complexity as \lname, but the core for MC-TFM++ is Transformer++ instead of GAU.

Figure~\ref{fig:model_ablation} shows that \lname outperforms MC-TFM++ by a large margin (more than 2$\times$ speedup when the sequence length is greater than 2048), confirming the importance of GAU in our design.
We further look into the perplexity increase due to our approximation method in Table \ref{tab:gap},
showing that the quality loss due to approximation is substantially smaller when going from \qname to \lname than going from TFM++ to MC-TFM++.
This indicates that \attn is more compatible with GAU than MHSA, which matches our intuition that GAU is more beneficial to weaker/approximate attention mechanisms.
\begin{table}[!ht]
    \centering
    \small
    \caption{\small Perplexity increases when \attn is applied to GAU ($\rightarrow$ FLASH) or to TFM++ ($\rightarrow$ MC-TFM++) --- \small{
    Results are reported for MLM and LM with increasing context lengths from 512 to 8192.}}
    \resizebox{\linewidth}{!}{
    \begin{tabular}{@{}c|ccccc@{}}
        \toprule
        MLM on C4 & 512 & 1024 & 2048 & 4096 & 8192 \\\midrule
        \qname $\rightarrow$ \lname & \bf 0.0  & \bf 0.05 & \bf 0.06 & \bf 0.07  & \bf 0.07  \\
        TFM++ $\rightarrow$ MC-TFM++ & 0.36 & 0.37 & 0.49 & 0.48 & 0.43 \\
        \midrule
        LM on Wiki-40B & 512 & 1024 & 2048 & 4096 & 8192 \\\midrule
        \qname $\rightarrow$ \lname & \bf -0.05 & \bf 0.06 & \bf 0.22 & \bf 0.30 & \bf 0.11 \\
        TFM++  $\rightarrow$ MC-TFM++ & 0.54 & 0.75 & 0.86 & 0.90 & 0.87 \\ 
        \bottomrule
    \end{tabular}
    }
    \label{tab:gap}
\end{table}

\paragraph{Impact of Chunk Size.} The choice of chunk size can affect both the quality and the training cost of \lname.
We observe that, in general, larger chunk sizes perform better as the context length increases.
For example, setting the chunk size to 512 is clearly preferable to the default chunk size (C=256) when the context length exceeds 1024.
In practice, hyperparameter search over the chunk size can be performed to optimize the performance of \lname further, although we did not explore such option in our experiments.
More detailed analysis can be found in Appendix~\ref{appendix:chunk_size}.

\section{Conclusion}
We have presented \name, a practical solution to address the quality  and empirical speed issues of existing efficient Transformer variants. This is achieved by designing a performant layer (gated linear unit) and by combining it with an accelerator-efficient approximation strategy (\attn). Experiments on bidirectional and auto-regressive language modeling tasks show that \name is as good as fully-augmented Transformers in quality (perplexity), while being substantially faster to train than the state-of-the-art.
A future work is to investigate the scaling laws of this new model family and the performance on downstream tasks.

\section*{Acknowledgements}
The authors would like to thank Gabriel Bender, John Blitzer, Maarten Bosma, Andrew Brock, Ed Chi, Hanjun Dai, Yann N. Dauphin, Pieter-Jan Kindermans and David So for their useful feedback.
Weizhe Hua was supported in part by the Facebook fellowship.

\bibliographystyle{icml2022}
\bibliography{main}

\newpage
\appendix
\onecolumn

\section{Connections to Combiner}
\label{sec:combiner}
To capture long-term information, Combiner~\cite{ren2021combiner} additionally summarizes each chunk into summary key and value vectors $K^\text{sum}, V^\text{sum} \in \mathbb{R}^{T/C \times d}$ and concatenate them into the local quadratic attention, i.e.
\begin{align*}
\hat{V}_g = \mathrm{Softmax}\big( Q [K_g ; K^\text{sum}] \big) [V_g ; V^\text{sum}].
\end{align*}
Effectively, Combiner compresses each chunk of $C$ vectors into a single vector of $\mathcal{O}(d)$,
whereas our chunked linear attention part compresses each chunk into a matrix ${K_h^\text{lin}}^\top V_h$ of size $\mathcal{O}(sd)$ which is $s$ times larger.
In other words, less compression is done in chunked linear attention,
allowing increased memory hence a potential advantage over Combiners.

Another difference lies in how the compressed long-term information from different chunks are combined, where Combiner reuses the quadratic attention whereas our chunked linear attention simply performs (cumulative) sum.
However, it is straightforward to incorporate what Combiner does in our proposed method by constructing an extra $[T/C \times T/C]$ attention matrix to combine the chunk summaries, e.g.
\begin{align*}
    A^\text{lin} &= \mathrm{relu}^2\Big( Q^\text{sum} {K^\text{sum}}^\top + b^\text{sum} \Big), \\
    \hat{V}_g^\text{lin} &= Q^\text{lin}_g \bigg[ \sum_{h=1}^{T/C} a^\text{lin}_{gh} \left( {K^\text{lin}_h}^\top V_h \right) \bigg].
\end{align*}
We indeed briefly experimented with this variant and found it helpful.
But it clearly complicates the overall model design, and more importantly requires the model to store and attend to all chunk summaries.
As a result, the auto-regressive decoding complexity will increase to $\mathcal{O}((C + T/C) d^2)$ which is length-dependent and no longer constant.
Hence, we do not include this feature in our default configuration.

\section{Experimental Setup }
\label{appendix:setup}

\subsection{Hyperparameters}
\paragraph{Bidirectional Language Modeling.}
Hyperparameters for the MLM task on C4 are listed in Table~\ref{tab:mlm_setting}. All models are implemented, trained, and evaluated using the same codebase to guarantee fair comparison.
\begin{table}[!ht]
\centering
\small
\caption{Hyperparameters for MLM pretraining on C4.}
\begin{threeparttable}
\begin{tabular}{@{}l|c@{}}
\toprule
 &  MLM Results (Figure~\ref{fig:mlm}) \\ \midrule
Data & C4 \\
Sequence length & 512 - 8192 \\
Tokens per batch & $2^{18}$ \\
Batch size &  $2^{18} / $ Sequence length\\
Number of steps & 125K \\
Warmup steps & 10K \\
Peak learning rate & 7e-4  \\
Learning rate decay & Linear \\
Optimizer & AdamW \\
Adam $\epsilon$ & 1e-6 \\
Adam $(\beta_1, \beta_2)$ & (0.9, 0.999) \\
Weight decay & 0.01 \\ 
Local gradient clipping* & 0.1 \\
Chunk size & 256 \\
Hidden dropout & 0 \\
GELU dropout & 0 \\
Attention dropout (if applicable) & 0 \\
\bottomrule
\end{tabular}
\begin{tablenotes}
  \item[*] \small{Applied to all models except the vanilla Transformer.}
\end{tablenotes}
\end{threeparttable}
\label{tab:mlm_setting}
\end{table}

\paragraph{Auto-regressive Language Modeling.}
Hyperparameters for the LM tasks on Wiki-40B and PG-19 are listed in Table~\ref{tab:lm_setting}.
All models are implemented, trained, and evaluated using the same codebase to guarantee fair comparison.
\begin{table}[!ht]
\centering
\small
\caption{Hyperparameters for LM pretraining on Wiki-40B and PG-19.}
\begin{threeparttable}
\begin{tabular}{@{}l|c|c@{}}
\toprule
 &  LM Results (Figure~\ref{fig:lm}) & LM Results (Table~\ref{tbl:pg19})\\ \midrule
Data & Wiki-40B & PG-19\\
Sequence length & 512 - 8192 & 1024 - 8192 \\ \cmidrule{2-3}
Tokens per batch & \multicolumn{2}{c}{$2^{18}$} \\
Batch size &  \multicolumn{2}{c}{$2^{18} / $ Sequence length} \\
Number of steps & \multicolumn{2}{c}{125K} \\
Warmup steps & \multicolumn{2}{c}{10K} \\
Peak learning rate & \multicolumn{2}{c}{7e-4} \\
Learning rate decay & \multicolumn{2}{c}{Linear} \\
Optimizer & \multicolumn{2}{c}{AdamW} \\
Adam $\epsilon$ & \multicolumn{2}{c}{1e-6} \\
Adam $(\beta_1, \beta_2)$ & \multicolumn{2}{c}{(0.9, 0.999)} \\
Weight decay & \multicolumn{2}{c}{0.01} \\ 
Local gradient clipping* & \multicolumn{2}{c}{0.1} \\
Hidden dropout &  \multicolumn{2}{c}{0} \\
GELU dropout &  \multicolumn{2}{c}{0} \\
Attention dropout (if applicable) &  \multicolumn{2}{c}{0} \\
\midrule
Chunk size & 256 & 512 \\
\bottomrule
\end{tabular}
\begin{tablenotes}
  \item[*] \small{Applied to all models except the vanilla Transformer.}
\end{tablenotes}
\end{threeparttable}
\label{tab:lm_setting}
\end{table}

\subsection{Model Specifications}
Detailed specifications of all models used in our experiments are summarized in Tables~\ref{tab:model_spec_c4},~\ref{tab:model_spec_wiki}, and~\ref{tab:model_spec_pg19}.
In the experiments, SiLU/Swish~\cite{elfwing2018sigmoid,hendrycks2016gelu,Ramachandran2017swish} is used as the nonlinearity for \qname and \lname, as it slightly outperforms GELU~\cite{hendrycks2016gelu} in our models.
It is also worth noting that we use ScaleNorm for some masked language models because ScaleNorm runs slightly faster than LayerNorm on TPU-v4 without compromising the quality of the model.

\begin{table}[!ht]
\centering
\caption{Model configurations for MLM experiments on the C4 dataset in Section \ref{sec:experiment}.}
\begin{threeparttable}
\resizebox{\textwidth}{!}{
\begin{tabular}{@{}l|ccccccc@{}}
\toprule
 & \qname & \lname & Transformer & Transformer+ & Transformer++ & Combiner & Performer \\ \midrule
\# of attention heads & 1 & 1 & 12 & 12 & 12 & 12 & 12 \\[0.2em]
Attention kernel & relu$^2$ & relu$^2$ & softmax & softmax & softmax & softmax & relu \\[0.2em]
Attention type & Quadratic & Mixed Chunk & Quadratic & Quadratic & Quadratic & Rowmajor-Axial & Linear \\[0.2em]
FFN type & GAU\textcolor{purple}{$^{1}$} & GAU\textcolor{purple}{$^{1}$}& MLP & MLP & GLU & MLP & MLP \\[0.2em]
Activation\textcolor{purple}{$^{2}$} & SiLU/Swish & SiLU/Swish & GELU & GELU & GELU & GELU & GELU \\[0.2em]
Norm. type\textcolor{purple}{$^{3}$} &  ScaleNorm & ScaleNorm & LayerNorm & ScaleNorm & ScaleNorm & ScaleNorm & ScaleNorm\\[0.2em]
Absolute position emb. & ScaledSin\textcolor{purple}{$^{4}$} & ScaledSin\textcolor{purple}{$^{4}$} & Learnable\textcolor{purple}{$^{5}$} & ScaledSin\textcolor{purple}{$^{4}$} & ScaledSin\textcolor{purple}{$^{4}$} & ScaledSin\textcolor{purple}{$^{4}$} & ScaledSin\textcolor{purple}{$^{4}$}\\[0.2em]
Relative position emb. & RoPE & RoPE & -- & RoPE & RoPE & RoPE & RoPE \\[0.2em]
\# of layers & 24 & 24 & 12+12\textcolor{purple}{$^{6}$} & 12+12\textcolor{purple}{$^{6}$} & 12+12\textcolor{purple}{$^{6}$} & 12+12\textcolor{purple}{$^{6}$} & 12+12\textcolor{purple}{$^{6}$} \\[0.2em]
Hidden size & 768 & 768 & 768 & 768 & 768 & 768 & 768 \\[0.2em]
Expansion rate &  2 & 2 & 4 & 4 & 4 & 4 & 4 \\[0.2em]
Chunk size & -- & 256 & -- & -- & -- & 256 & --  \\[0.2em]
Params (M) & 112 & 112 & 110 & 110 & 110 & 124 & 110\\
\bottomrule
\end{tabular}}
\begin{tablenotes}
  \item[\textcolor{purple}{1}] \small{\qname and \lname combines the attention and feed-forward network into one module named GAU.}
  \item[\textcolor{purple}{2}] \small{SiLU/Swish are proposed by~\citet{elfwing2018sigmoid,hendrycks2016gelu,Ramachandran2017swish}.}
  \item[\textcolor{purple}{3}] \small{ScaleNorm and LayerNorm are proposed by~\citet{Nguyen2019scalenorm} and~\citet{ba2016layer}, respectively.}
  \item[\textcolor{purple}{4}] \small{ScaleSin re-scales sinusoidal position embedding~\cite{vaswani2017attention} with a linearnable scalar for stability.}
  \item[\textcolor{purple}{5}] \small{The learnable position embedding is proposed by~\citet{Jonas2017convseq2seq}.}
  \item[\textcolor{purple}{6}] \small{The model is consist of 12 attention layers and 12 FFN layers.}
\end{tablenotes}
\end{threeparttable}
\label{tab:model_spec_c4}
\end{table}

\begin{table}[!ht]
\centering
\caption{Model configurations for LM experiments on the Wiki-40B dataset in Section \ref{sec:experiment}.}
\begin{threeparttable}
\resizebox{\textwidth}{!}{
\begin{tabular}{@{}l|ccccccc@{}}
\toprule
 & \qname & \lname & Transformer & Transformer+ & Transformer++ & Combiner & Performer \\ \midrule
\# of attention heads & 1 & 1 & 12 & 12 & 12 & 12 & 12 \\[0.2em]
Attention kernel & relu$^2$ & relu$^2$ & softmax & softmax & softmax & softmax & relu \\[0.2em]
Attention type & Quadratic & Mixed Chunk & Quadratic & Quadratic & Quadratic & Rowmajor-Axial & Linear \\[0.2em]
FFN type & GAU\textcolor{purple}{$^{1}$} & GAU\textcolor{purple}{$^{1}$} & MLP & MLP & GLU & MLP & MLP \\[0.2em]
Activation\textcolor{purple}{$^{2}$} &  SiLU/Swish & SiLU/Swish & GELU & GELU & GELU & GELU & GELU \\[0.2em]
Norm. type & LayerNorm & LayerNorm & LayerNorm & LayerNorm & LayerNorm & LayerNorm & LayerNorm \\[0.2em]
Absolute position emb. & ScaledSin\textcolor{purple}{$^{3}$} & ScaledSin\textcolor{purple}{$^{3}$} & Learnable\textcolor{purple}{$^{4}$} & ScaledSin\textcolor{purple}{$^{3}$} & ScaledSin\textcolor{purple}{$^{3}$} & ScaledSin\textcolor{purple}{$^{3}$} & ScaledSin\textcolor{purple}{$^{3}$}\\[0.2em]
Relative position emb. & RoPE & RoPE & -- & RoPE & RoPE & RoPE & RoPE \\[0.2em]
\# of layers & 24 & 24 & 12+12\textcolor{purple}{$^{5}$} & 12+12\textcolor{purple}{$^{5}$} & 12+12\textcolor{purple}{$^{5}$} & 12+12\textcolor{purple}{$^{5}$} & 12+12\textcolor{purple}{$^{5}$} \\[0.2em]
Hidden size & 768 & 768 & 768 & 768 & 768 & 768 & 768 \\[0.2em]
Expansion rate &  2 & 2 & 4 & 4 & 4 & 4 & 4 \\[0.2em]
Chunk size & -- & 256 & -- & -- & -- & 256 & --  \\[0.2em]
Params (M) & 112 & 112 & 110 & 110 & 110 & 124 & 110\\
\bottomrule
\end{tabular}}
\begin{tablenotes}
  \item[\textcolor{purple}{1}] \small{\qname and \lname combines the attention and feed-forward network into one module named GAU.}
  \item[\textcolor{purple}{2}] \small{SiLU/Swish are proposed by~\citet{elfwing2018sigmoid,hendrycks2016gelu,Ramachandran2017swish}.}
  \item[\textcolor{purple}{3}] \small{ScaleSin re-scales sinusoidal position embedding~\cite{vaswani2017attention} with a linearnable scalar for stability.}
  \item[\textcolor{purple}{4}] \small{The learnable position embedding is proposed by~\citet{Jonas2017convseq2seq}.}
  \item[\textcolor{purple}{5}] \small{The model is consist of 12 attention layers and 12 FFN layers.}
\end{tablenotes}
\end{threeparttable}
\label{tab:model_spec_wiki}
\end{table}

\begin{table*}[!ht]
\centering
\caption{Model configurations for LM experiments on the PG-19 dataset in Section \ref{sec:experiment}.}
\begin{threeparttable}
\resizebox{0.8\textwidth}{!}{
\begin{tabular}{@{}l|cccccc@{}}
\toprule
 & \qname & \lname & Transformer+ & Transformer++ & Combiner \\ \midrule
\# of attention heads & 1 & 1 & 16 & 16 & 16 \\[0.2em]
Attention kernel & relu$^2$ & relu$^2$ & softmax & softmax & softmax \\[0.2em]
Attention type & Quadratic & Mixed Chunk & Quadratic & Quadratic & Rowmajor-Axial  \\[0.2em]
FFN type & GAU\textcolor{purple}{$^{1}$}& GAU\textcolor{purple}{$^{1}$} & MLP & GLU & MLP  \\[0.2em]
Activation\textcolor{purple}{$^{2}$} &  SiLU/Swish & SiLU/Swish & GELU & GELU & GELU \\[0.2em]
Norm. type & LayerNorm & LayerNorm & LayerNorm & LayerNorm & LayerNorm\\[0.2em]
Absolute position emb. & ScaledSin\textcolor{purple}{$^{3}$} & ScaledSin\textcolor{purple}{$^{3}$} & ScaledSin\textcolor{purple}{$^{3}$} & ScaledSin\textcolor{purple}{$^{3}$} & ScaledSin\textcolor{purple}{$^{3}$}\\[0.2em]
Relative position emb. & RoPE & RoPE & RoPE & RoPE & RoPE \\[0.2em]
\# of layers & 72 & 72 & 36+36\textcolor{purple}{$^{4}$} & 36+36\textcolor{purple}{$^{4}$} & 36+36\textcolor{purple}{$^{4}$}  \\[0.2em]
Hidden size & 1024 & 1024 & 1024 & 1024 & 1024 \\[0.2em]
Expansion rate &  2 & 2 & 4 & 4 & 4   \\[0.2em]
Chunk size & -- & 512 & -- &  -- & 512   \\[0.2em]
Params (M) & 496 & 496 & 486 & 486 & 562 \\
\bottomrule
\end{tabular}}
\begin{tablenotes}
  \item[\textcolor{purple}{1}] \small{\qname and \lname combines the attention and feed-forward network into one module named GAU.}
  \item[\textcolor{purple}{2}] \small{SiLU/Swish are proposed by~\citet{elfwing2018sigmoid,hendrycks2016gelu,Ramachandran2017swish}.}
  \item[\textcolor{purple}{3}] \small{ScaleSin re-scales sinusoidal position embedding~\cite{vaswani2017attention} with a linearnable scalar for stability.}
  \item[\textcolor{purple}{4}] \small{The model is consist of 36 attention layers and 36 FFN layers.}
\end{tablenotes}
\end{threeparttable}
\label{tab:model_spec_pg19}
\end{table*}

\section{Additional Experimental Results}
\label{appendix:more_results}
Here, we provide full results on the training speed of different language models using a Nvidia V100 GPU (in Table~\ref{tab:gpu_speed}) and the ablation study of chunk size for \lname (in Figure~\ref{fig:chunk_size}).

\begin{table*}[!ht]
\centering
\small
\caption{Comparison of latency for each training step of auto-regressive language modeling on Wiki-40B using a single Nvidia Tesla V100 GPU --- \small{Latency is reported in millisecond. OOM stands for the CUDA out of memory error. Performer-Matmul implements the cumulative sum (\texttt{cumsum}) using matrix multiplication}.}
\begin{threeparttable}
\begin{tabular}{@{}lcccc@{}}
\toprule
 & \multicolumn{4}{c}{Context length $\times$ Batch size} \\ \cmidrule(l){2-5} 
Model & 512 $\times$ 4 & 1024 $\times$ 2 & 2048 $\times$ 1 & 4096 $\times$ 1\\ \midrule
Transformer++ & \textbf{222.4} & 243.9 & 315.0 & OOM\\
Performer & 823.0 & 827.4 & 799.8 & OOM\\
Performer-Matmul & 697.4 & 701.7 & 688.9 & OOM\\ \midrule
\lname & 254.4 & \textbf{235.0} & \textbf{242.8} & \textbf{452.9} \\ \bottomrule
\end{tabular}
\end{threeparttable}
\label{tab:gpu_speed}
\end{table*}

\subsection{Auto-regressive Training on GPU}
\label{appendix:gpu_training}
We observe that the inefficiency of auto-regressive training is not limited to hardware accelerators such as TPUs.
As shown in Table~\ref{tab:gpu_speed}, Performer has the largest latency among the three models because it requires to perform \texttt{cumsum} over all tokens sequentially.
In contrast, the proposed \lname achieves the lowest latency when the context length is over 1024, suggesting the effectiveness of the proposed \attn mechanism.

\subsection{Tabular MLM and LM Results}
\label{appendix:tabular_results}
We summarize the experimental results of MLM on C4 and LM on Wiki-40B in Tables~\ref{tab:mlm_c4} and~\ref{tab:lm_wiki}.
\begin{table}[!ht]
\caption{Bidirectional/masked language models on the C4 dataset. The best perplexity (PPLX) on the validation set is reported. Training latency is measured with 64 TPU-v4 cores.}
\label{tab:mlm_c4}
\centering
\small
\begin{threeparttable}
\begin{tabular}{@{}l|cccccccccc@{}}
\toprule
\multirow{3}{*}{Model} & \multicolumn{10}{c}{Context Length} \\ \cmidrule(l){2-11} 
 & \multicolumn{2}{c|}{512} & \multicolumn{2}{c|}{1024} & \multicolumn{2}{c|}{2048} & \multicolumn{2}{c|}{4096} & \multicolumn{2}{c}{8192} \\ \cmidrule(l){2-11} 
 & PPLX & \multicolumn{1}{c|}{Latency} & PPLX & \multicolumn{1}{c|}{Latency} & PPLX & \multicolumn{1}{c|}{Latency} & PPLX & \multicolumn{1}{c|}{Latency} & PPLX & Latency \\ \midrule
\multicolumn{1}{l|}{Transformer} & 4.517 & \multicolumn{1}{c|}{47.7} & 4.436 & \multicolumn{1}{c|}{63.9} & 4.196 & \multicolumn{1}{c|}{90.9} & 4.602 & \multicolumn{1}{c|}{142.5} & 4.8766 & 252.7 \\
\multicolumn{1}{l|}{Transformer+} & 4.283 & \multicolumn{1}{c|}{48.8} & 4.151 & \multicolumn{1}{c|}{64.4} & 4.032 & \multicolumn{1}{c|}{91.5} & 3.989 & \multicolumn{1}{c|}{142.9} & 3.986 & 252.9\\
\multicolumn{1}{l|}{Transformer++} & 4.205 & \multicolumn{1}{c|}{47.6} & 4.058 & \multicolumn{1}{c|}{64.6} & 3.920 & \multicolumn{1}{c|}{91.6} & 3.876 & \multicolumn{1}{c|}{143.4} & 3.933 & 252.1\\ \midrule
\multicolumn{1}{l|}{Performer} & 5.897 & \multicolumn{1}{c|}{\textbf{37.2}} & 6.324 & \multicolumn{1}{c|}{\textbf{37.6}} & 8.032 & \multicolumn{1}{c|}{\textbf{39.1}} & 12.622 & \multicolumn{1}{c|}{\textbf{36.9}} & 102.980 & \textbf{40.9}\\
\multicolumn{1}{l|}{Combiner} & 4.449 & \multicolumn{1}{c|}{67.2} & 4.317 & \multicolumn{1}{c|}{66.4} & 4.238 & \multicolumn{1}{c|}{66.4} & 4.195 & \multicolumn{1}{c|}{68.3} & 4.225 & 77.3 \\ \midrule
\multicolumn{1}{l|}{\qname} & \textbf{4.176} & \multicolumn{1}{c|}{43.7} & \textbf{3.964} & \multicolumn{1}{c|}{50.1} & \textbf{3.864} & \multicolumn{1}{c|}{61.7} & \textbf{3.828} & \multicolumn{1}{c|}{84.9} & \textbf{3.830} & 132.1\\
\multicolumn{1}{l|}{\lname} & \textbf{4.172} & \multicolumn{1}{c|}{51.2} & 4.015 & \multicolumn{1}{c|}{50.1} & 3.928 & \multicolumn{1}{c|}{51.4} & 3.902 & \multicolumn{1}{c|}{50.7} & 3.897 & 59.9\\ \bottomrule
\end{tabular}
\end{threeparttable}
\end{table}

\begin{table}[!ht]
\caption{Auto-regressive language models on the Wiki-40B dataset. The best perplexity (PPLX) on the validation set is reported. Training latency is measured with 64 TPU-v4 cores.}
\label{tab:lm_wiki}
\centering
\small
\begin{threeparttable}
\begin{tabular}{@{}l|cccccccccc@{}}
\toprule
\multirow{3}{*}{Model} & \multicolumn{10}{c}{Context Length} \\ \cmidrule(l){2-11} 
 & \multicolumn{2}{c|}{512} & \multicolumn{2}{c|}{1024} & \multicolumn{2}{c|}{2048} & \multicolumn{2}{c|}{4096} & \multicolumn{2}{c}{8192} \\ \cmidrule(l){2-11} 
 & PPLX & \multicolumn{1}{c|}{Latency} & PPLX & \multicolumn{1}{c|}{Latency} & PPLX & \multicolumn{1}{c|}{Latency} & PPLX & \multicolumn{1}{c|}{Latency} & PPLX & Latency \\ \midrule
\multicolumn{1}{l|}{Transformer} & 17.341 & \multicolumn{1}{c|}{\textbf{54.0}} & 19.808 & \multicolumn{1}{c|}{70.9} & 18.154 & \multicolumn{1}{c|}{96.3} & 17.731 & \multicolumn{1}{c|}{149.1} & 18.254 & 260.7 \\
\multicolumn{1}{l|}{Transformer+} & 16.907 & \multicolumn{1}{c|}{55.6} & 15.999 & \multicolumn{1}{c|}{70.3} & 15.653 & \multicolumn{1}{c|}{96.1} & 15.515 & \multicolumn{1}{c|}{149.3} & 15.478 & 261.9\\
\multicolumn{1}{l|}{Transformer++} & 16.835 & \multicolumn{1}{c|}{54.7} & 15.943 & \multicolumn{1}{c|}{70.9} & 15.489 & \multicolumn{1}{c|}{96.6} & 15.282 & \multicolumn{1}{c|}{149.2} & 15.254 & 261.0\\ \midrule
\multicolumn{1}{l|}{Performer} & 18.989 & \multicolumn{1}{c|}{1439.7} & 18.520 & \multicolumn{1}{c|}{1386.9} & 18.547 & \multicolumn{1}{c|}{1518.9} & 18.987 & \multicolumn{1}{c|}{1526.7} & 19.923 & 1526.8\\
\multicolumn{1}{l|}{Combiner} & 17.338 & \multicolumn{1}{c|}{75.5} & 16.710 & \multicolumn{1}{c|}{74.4} & 16.344 & \multicolumn{1}{c|}{71.8} & 16.171 & \multicolumn{1}{c|}{71.7} & 16.119 & 77.9 \\ \midrule
\multicolumn{1}{l|}{\qname} & 16.633 & \multicolumn{1}{c|}{\textbf{54.1}} & \textbf{15.879} & \multicolumn{1}{c|}{59.5} & \textbf{15.305} & \multicolumn{1}{c|}{71.3} & \textbf{14.955} & \multicolumn{1}{c|}{96.1} & \textbf{14.998} & 141.3\\
\multicolumn{1}{l|}{\lname} & \textbf{16.581} & \multicolumn{1}{c|}{57.2} & 15.935 & \multicolumn{1}{c|}{\textbf{56.9}} & 15.525 & \multicolumn{1}{c|}{\textbf{56.7}} & 15.259 & \multicolumn{1}{c|}{\textbf{57.0}} & 15.109 & \textbf{62.5}\\ \bottomrule
\end{tabular}
\end{threeparttable}
\end{table}

\subsection{Ablation Study of Chunk Size}
\label{appendix:chunk_size}
\begin{figure*}[!ht]
\centering
\includegraphics[width=0.99\textwidth]{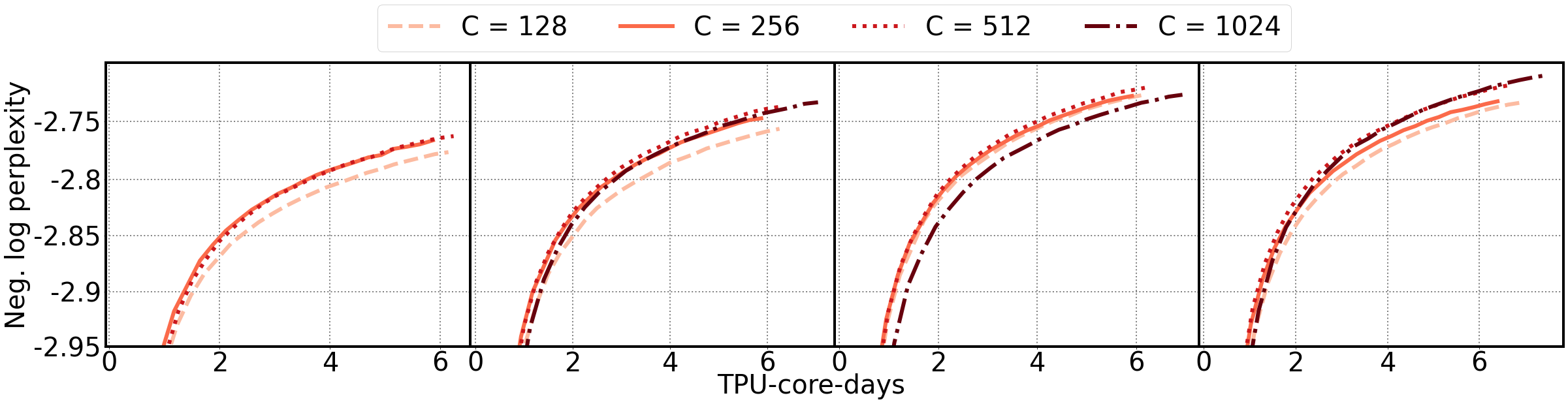}
\hspace*{0.3cm}\small{\hfil{(a) Context length = 1024}}
\small{\hfil{(b) Context length = 2048}}
\small{\hfil{(c) Context length = 4096}}
\small{\hfil{(d) Context length = 8192}}
\caption{Ablation study of the chunk size (C) of \lname for context lengths from 1K to 8K. 
}
\label{fig:chunk_size}
\end{figure*}

The choice of chunk size can have an impact on both the quality and the training cost of \lname.
In the extreme case where chunk size equals the context length, \lname falls back to \qname and loses the scalability to long context lengths.
In the other extreme case where chunk size is equal to one, the proposed attention module becomes a linear attention, which suffers from inefficient auto-regressive training.
Figure~\ref{fig:chunk_size} shows the tradeoff between the quality and training cost of four different chunk sizes  for context lengths from 1K to 8K.

\section{Pseudocode For \qname and \lname}
We show the detailed implementation of \qname and \lname in Codes~\ref{code:flashq} and~\ref{code:flash}.

\begin{figure}[!ht]
\centering
\begin{minipage}{0.75\linewidth}
\begin{lstlisting}[language=python, mathescape]
def _get_scaledsin(embeddings):
    """Create sinusoidal position embedding with a scaling factor."""
    hidden_size = int(embeddings.shape[-1])
    pos = tf.range(tf.shape(embeddings)[1])
    pos = tf.cast(pos, tf.float32)
    half_d = hidden_size // 2
    freq_seq = tf.cast(tf.range(half_d), tf.float32) / float(half_d)
    inv_freq = 10000 ** -freq_seq
    sinusoid = tf.einsum('s,d$\rightarrow$sd', pos, inv_freq)
    scaledsin = tf.concat([tf.sin(sinusoid), tf.cos(sinusoid)], axis=-1)
    scalar = tf.get_variable(
        'scaledsin_scalar',
        shape=(),
        initializer=tf.constant_initializer(1 / hidden_size ** 0.5))
    scaledsin *= scalar
    return scaledsin
\end{lstlisting}
\captionof{Code}{Pseudocode for ScaledSin absolute position embedding.}
\label{code:scaledsin}
\end{minipage}
\end{figure}

\begin{figure}[!ht]
\centering
\begin{minipage}{0.75\linewidth}
\begin{lstlisting}[language=python, mathescape]
def rope(x, axis):
    """RoPE position embedding."""
    shape = x.shape.as_list()
    if isinstance(axis, int):
        axis = [axis]

    spatial_shape = [shape[i] for i in axis]
    total_len = 1
    for i in spatial_shape:
        total_len *= i
    position = tf.reshape(
        tf.cast(tf.range(total_len, delta=1.0), tf.float32), spatial_shape)
        
    for i in range(axis[-1] + 1, len(shape) - 1, 1):
        position = tf.expand_dims(position, axis=-1)
        
    half_size = shape[-1] // 2
    freq_seq = tf.cast(tf.range(half_size), tf.float32)/float(half_size)
    inv_freq = 10000 ** -freq_seq
    sinusoid = tf.einsum('...,d$\rightarrow$...d', position, inv_freq)
    sin = tf.sin(sinusoid)
    cos = tf.cos(sinusoid)
    x1, x2 = tf.split(x, 2, axis=-1)
    return tf.concat([x1 * cos - x2 * sin, x2 * cos + x1 * sin], axis=-1)
\end{lstlisting}
\captionof{Code}{Pseudocode for RoPE.}
\label{code:rope}
\end{minipage}
\end{figure}

\begin{figure}[!ht]
\centering
\begin{minipage}{0.75\linewidth}
\begin{lstlisting}[language=python, mathescape]
WEIGHT_INITIALIZER = tf.random_normal_initializer(stddev=0.02)

def rel_pos_bias(n):
    """Relative position bias."""
    if n < 512:
    # Construct Toeplitz matrix directly when the sequence length is less than 512.
        w = tf.get_variable(
            'weight',
            shape=[2 * n - 1],
            dtype=tf.float32,
            initializer=WEIGHT_INITIALIZER)
        t = tf.pad(w, [[0, n]])
        t = tf.tile(t, [n])
        t = t[..., :-n]
        t = tf.reshape(t, [n, 3 * n - 2])
        r = (2 * n - 1) // 2
        t = t[..., r:-r]
    else:
    # Construct Toeplitz matrix using RoPE when the sequence length is over 512.
        a = tf.get_variable(
            'a',
            shape=[128],
            dtype=dtype,
            initializer=WEIGHT_INITIALIZER)
        b = tf.get_variable(
            'b',
            shape=[128],
            dtype=dtype,
            initializer=WEIGHT_INITIALIZER)
        a = rope(tf.tile(a[None, :], [n, 1]), axis=0)
        b = rope(tf.tile(b[None, :], [n, 1]), axis=0)
        t = tf.einsum('mk,nk$\rightarrow$mn', a, b)
return t
\end{lstlisting}
\captionof{Code}{Pseudocode for relative position bias.}
\label{code:rel_pos_bias}
\end{minipage}
\end{figure}

\begin{figure}[!ht]
\centering
\begin{minipage}{0.75\linewidth}
\begin{lstlisting}[language=python, mathescape]
def norm(x, begin_axis=-1, eps=1e-5, norm_type='layer_norm'):
    """Normalization layer."""
    shape = x.shape.as_list()
    axes = list(range(len(shape)))[begin_axis:]
    if norm_type == 'layer_norm':
        mean, var = tf.nn.moments(x, axes, keepdims=True)
        x = (x - mean) * tf.rsqrt(var + eps)
        gamma = tf.get_variable(
            'gamma', shape=x.shape.as_list()[begin_axis:], initializer=tf.initializers.ones())
        beta = tf.get_variable(
            'beta', shape=x.shape.as_list()[begin_axis:], initializer=tf.initializers.zeros())
        return gamma * x + beta
    elif norm_type == 'scale_norm':
        mean_square =tf.reduce_mean(tf.math.square(x), axes, keepdims=True)
        x = x * tf.rsqrt(mean_square + eps)
        scalar = tf.get_variable('scalar', shape=(), initializer=tf.constant_initializer(1.0))
        return scale * x
\end{lstlisting}
\captionof{Code}{Pseudocode for LayerNorm and ScaleNorm.}
\label{code:norm}
\end{minipage}
\end{figure}

\begin{figure}[!ht]
\centering
\begin{minipage}{0.75\linewidth}
\begin{lstlisting}[language=python,
mathescape]
WEIGHT_INITIALIZER = tf.random_normal_initializer(stddev=0.02)

def GAU(x, causal, norm_type='layer_norm', expansion_factor=2):
    """GAU block.
    
    Input shape: batch size x sequence length x model size
    """
    seq_len = tf.shape(x)[1]
    d = int(x.shape[-1])
    e = int(d * expansion_factor)
    
    shortcut, x = x, norm(x, begin_axis=-1, norm_type=norm_type)
    
    s = 128
    uv = tf.layers.dense(x, 2 * e + s, kernel_initializer=WEIGHT_INITIALIZER, bias_initializer='zeros')
    u, v, base = tf.split(tf.nn.silu(uv), [e, e, s], axis=-1)
    
    # Generate Query (q) and Key (k) from base.
    gamma = tf.get_variable('gamma', shape=[2, s], initializer=WEIGHT_INITIALIZER)
    beta = tf.get_variable('beta', shape=[2, s], initializer=tf.initializers.zeros())
    base = tf.einsum('...r,hr$\rightarrow$...hr', base, gamma) + beta
    base = rope(base, axis=1)
    q, k = tf.unstack(base, axis=-2)
    
    # Calculate the quadratic attention.
    qk = tf.einsum('bnd,bmd$\rightarrow$bnm', q, k)
    bias = rel_pos_bias(seq_len)
    kernel = tf.math.square(tf.nn.relu(qk / seq_len + bias))
    
    # Apply the causal mask for auto-regressive tasks.
    if causal:
        causal_mask = tf.linalg.band_part(
            tf.ones([seq_len, seq_len], dtype=x.dtype), num_lower=-1, num_upper=0)
        kernel *= causal_mask
        
    x = u * tf.einsum('bnm,bme$\rightarrow$bne', kernel, v)
    x = tf.layers.dense(x, d, kernel_initializer=WEIGHT_INITIALIZER, bias_initializer='zeros')
    return x + shortcut
\end{lstlisting}
\captionof{Code}{Pseudocode for GAU (\qname).}
\label{code:flashq}
\end{minipage}
\end{figure}

\begin{figure}[!ht]
\centering
\begin{minipage}{0.75\linewidth}
\begin{lstlisting}[language=python,
mathescape]
def segment_ids_to_mask(segment_ids, causal=False):
    """Generate the segment mask from the segment ids. 
    
    The segment mask is used to remove the attention between tokens in different documents.
    """
    min_ids, max_ids = tf.reduce_min(segment_ids, axis=-1), tf.reduce_max(segment_ids, axis=-1)
    # 1.0 indicates in the same group and 0.0 otherwise
    mask = tf.logical_and(
        tf.less_equal(min_ids[:, :, None], max_ids[:, None, :]),
        tf.greater_equal(max_ids[:, :, None], min_ids[:, None, :]))
    mask = tf.cast(mask, tf.float32)
    if causal:
        g = tf.shape(min_ids)[1]
        causal_mask = 1.0 - tf.linalg.band_part(
            tf.ones([g, g], dtype=tf.float32), num_lower=0, num_upper=-1)
        mask *= causal_mask
    mask = tf.math.divide_no_nan(mask, tf.reduce_sum(mask, axis=-1, keepdims=True))
    return mask
\end{lstlisting}
\captionof{Code}{Pseudocode for generating segment mask.}
\label{code:segmask}
\end{minipage}
\end{figure}

\begin{figure}[!ht]
\centering
\begin{minipage}{0.75\linewidth}
\begin{lstlisting}[language=python,
mathescape]
WEIGHT_INITIALIZER = tf.random_normal_initializer(stddev=0.02)

def FLASH(x, causal, segment_ids, norm_type='layer_norm', expansion_factor=2):
    """FLASH block.
    
    Input shape: batch size x num chunks x chunk length x model size
    """
    _, g, n, d = x.shape.as_list()
    e = int(d * expansion_factor)
    shortcut, x = x, norm(x, begin_axis=-1, norm_type=norm_type)

    s = 128
    uv = tf.layers.dense(x, 2 * e + s, kernel_initializer=WEIGHT_INITIALIZER, bias_initializer='zeros')
    u, v, base = tf.split(tf.nn.silu(uv), [e, e, s], axis=-1)
    
    # Generate Query and Key for both quadratic and linear attentions.
    gamma = tf.get_variable('gamma', shape=[4, s], initializer=WEIGHT_INITIALIZER)
    beta = tf.get_variable('beta', shape=[4, s], initializer=tf.initializers.zeros())
    base = tf.einsum('...r,hr$\rightarrow$...hr', base, gamma) + beta
    base = rope(base, axis=[1, 2])
    quad_q, quad_k, lin_q, lin_k = tf.unstack(base, axis=-2)

    if causal:
        # Linear attention part.
        lin_kv = tf.einsum('bgnk,bgne$\rightarrow$bgke', lin_k, v) / tf.cast(n, x.dtype)
        mask = segment_ids_to_mask(segment_ids, causal=True)
        cum_lin_kv = tf.einsum('bhke,bgh$\rightarrow$bgke', lin_kv, mask)
        linear = tf.einsum('bgnk,bgke$\rightarrow$bgne', lin_q, cum_lin_kv)
        
        # Quadratic attention part.
        quad_qk = tf.einsum('bgnk,bgmk$\rightarrow$bgnm', quad_q, quad_k)
        bias = rel_pos_bias(n)
        kernel = tf.math.square(tf.nn.relu(quad_qk / n + bias))
        # Apply the causal mask for auto-regressive tasks.
        causal_mask = tf.linalg.band_part(tf.ones([n, n], dtype=x.dtype), num_lower=-1, num_upper=0)
        quadratic = tf.einsum('bgnm,bgme$\rightarrow$bgne', kernel * causal_mask, v)
    else:
        # Linear attention part
        lin_kv = tf.einsum('bgnk,bgne$\rightarrow$bgke', lin_k, v) / tf.cast(n, x.dtype)
        mask = segment_ids_to_mask(segment_ids)
        lin_kv = tf.einsum('bhke,bgh$\rightarrow$bgke', lin_kv, mask)
        linear = tf.einsum('bgnk,bgke$\rightarrow$bgne', lin_q, lin_kv)
        
        # Quadratic attention part
        quad_qk = tf.einsum('bgnk,bgmk$\rightarrow$bgnm', quad_q, quad_k)
        bias = rel_pos_bias(n)
        kernel = tf.math.square(tf.nn.relu((quad_qk / n + bias))
        quadratic = tf.einsum('bgnm,bgme$\rightarrow$bgne', kernel, v)

    x = u * (quadratic + linear)
    x = tf.layers.dense(x, d, kernel_initializer=WEIGHT_INITIALIZER, bias_initializer='zeros')
    return x + shortcut
\end{lstlisting}
\captionof{Code}{Pseudocode for \name.}
\label{code:flash}
\end{minipage}
\end{figure}

\end{document}